\pdfoutput=1

\documentclass[11pt]{article}

\usepackage[preprint]{acl}

\usepackage{times}
\usepackage{latexsym}

\usepackage[T1]{fontenc}

\usepackage[utf8]{inputenc}

\usepackage{microtype}

\usepackage{inconsolata}

\usepackage{graphicx}
\usepackage{kotex}
\usepackage{xcolor}
\usepackage{multirow}
\usepackage{makecell}
\usepackage{subcaption}
\usepackage{tcolorbox}
\usepackage{siunitx}
\usepackage{xspace}
\usepackage{geometry}
\usepackage{tabularx}
\usepackage{array}
\usepackage{lipsum} 
\usepackage{booktabs}
\usepackage{multirow}
\usepackage{soul}
\usepackage{tabularx}

\newcommand{\mymethod}[0]{JoA-ICL\xspace}
\newcommand{\mydata}[0]{K-News-Stance\xspace}

\title{Journalism-Guided Agentic In-Context Learning\\for News Stance Detection}

\author{
Dahyun Lee$^{\heartsuit}$~~~~~~Jonghyeon Choi$^{\clubsuit}$~~~~~~Jiyoung Han$^{\spadesuit}$~~~~~~Kunwoo Park$^{\diamondsuit\clubsuit}$\\ 
$^{\heartsuit}$Department of Metaverse and Culture Contents, Soongsil University\\
$^{\clubsuit}$Department of Intelligent Semiconductors, Soongsil University\\
$^{\spadesuit}$Graduate School of Future Strategy, KAIST\\
$^{\diamondsuit}$School of AI Convergence, Soongsil University\\
\texttt{kunwoo.park@ssu.ac.kr}, \texttt{jiyoung.han@kaist.ac.kr}
}

\begin{document}

\maketitle

\begin{abstract}
As online news consumption grows, personalized recommendation systems have become integral to digital journalism. However, these systems risk reinforcing filter bubbles and political polarization by failing to incorporate diverse perspectives. Stance detection---identifying a text's position on a target---can help mitigate this by enabling viewpoint-aware recommendations and data-driven analyses of media bias. Yet, existing stance detection research remains largely limited to short texts and high-resource languages. To address these gaps, we introduce \textsc{\mydata}, the first Korean dataset for article-level stance detection, comprising 2,000 news articles with article-level and 21,650 segment-level stance annotations across 47 societal issues. We also propose \textsc{\mymethod}, a \textbf{Jo}urnalism-guided \textbf{A}gentic \textbf{I}n-\textbf{C}ontext \textbf{L}earning framework that employs a language model agent to predict the stances of key structural segments (e.g., leads, quotations), which are then aggregated to infer the overall article stance. Experiments showed that \textsc{\mymethod} outperforms existing stance detection methods, highlighting the benefits of segment-level agency in capturing the overall position of long-form news articles. Two case studies further demonstrate its broader utility in promoting viewpoint diversity in news recommendations and uncovering patterns of media bias.
\end{abstract}

\section{Introduction}

With the proliferation of digital platforms, online news consumption has become ubiquitous. In response, major news providers have shifted their publication channels from offline newspapers to online newspapers~\cite{martens2018digital, 10.1257/app.20210689}, and adopted personalized recommendation algorithms to enhance the experience of news readers~\cite{feng2020news,10.1145/3530257}. However, such systems may inadvertently confine users within limited information environments, leading to filter bubbles and echo chambers that intensify political polarization~\cite{flaxman2016filter, 10.1145/3614419.3643996}. To mitigate these effects, it is essential to automatically identify the perspectives embedded in news content and integrate them into recommendation algorithms, thereby promoting a more balanced media ecosystem.

Stance detection is a natural language processing task that identifies the position expressed in a text toward a specific target~\cite{2020Kucuk, hardalov-etal-2022-survey}. Applied to news articles, stance detection can facilitate balanced recommendations by incorporating diverse viewpoints, thereby enabling users to make more informed decisions~\cite{alam-2022-towards,reuver-etal-2024-investigating}. It also provides a data-driven approach to examine media bias, allowing outlet-wise comparisons of stance distributions across diverse issues~\cite{kuila2024news}.

\begin{figure}[t]
\includegraphics[width=0.99\linewidth]{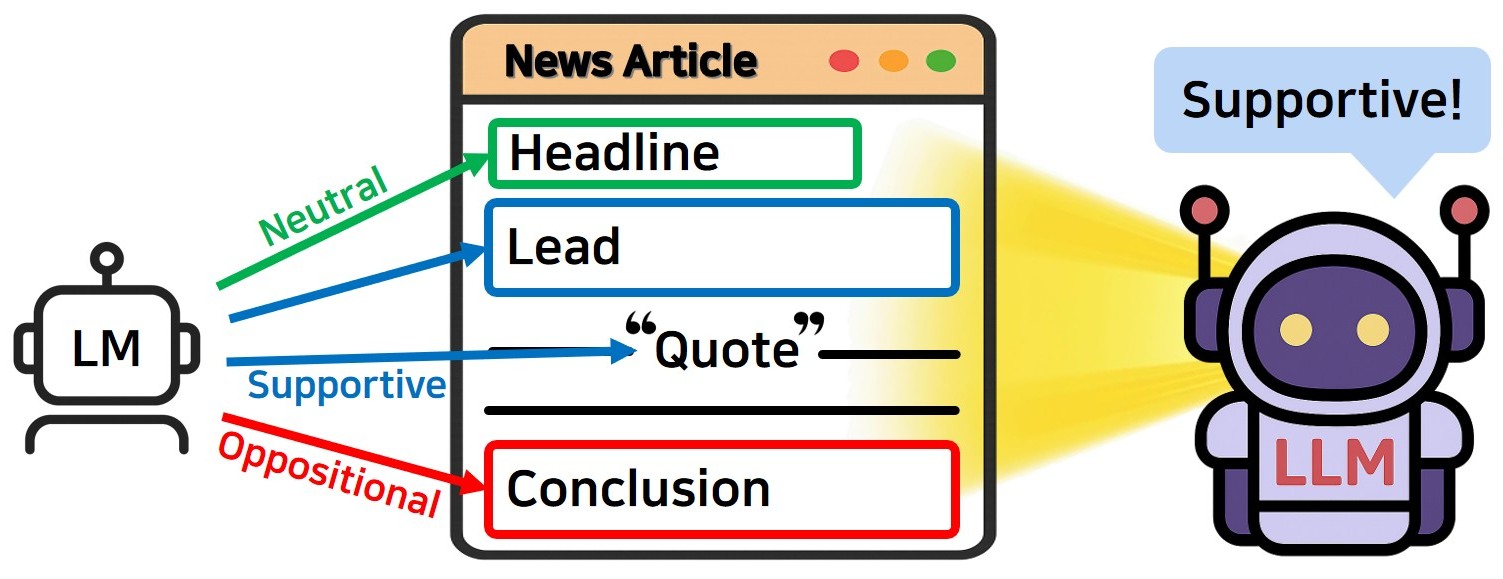}
\caption{Key idea of \textsc{\mymethod}, illustrating how article-level detection is performed by leveraging segment-level predictions generated by a language model agent.}
\label{fig:intro}
\end{figure}

Despite the growing need for stance detection methods in news articles, two significant gaps remain in prior research. First, most existing studies focus on short texts, such as individual sentences or tweets~\cite{darwish2020unsupervised, glandt-etal-2021-stance, evrard-etal-2020-french}. In contrast, news articles are often much longer, sometimes exceeding a thousand words. Within such lengthy texts, nuanced stances may vary across different segments. This makes it challenging for models to accurately infer the overall stance. Second, available datasets are mainly limited to high-resource languages~\cite{li-etal-2021-p, mascarell-etal-2021-stance}, such as English and German. The resource gap is even more significant for news article-level datasets. To enable more comprehensive and culturally grounded stance detection, it is crucial to develop datasets in non-major languages that reflect country-specific issues and linguistic nuances.

To address these gaps, this study introduces \textsc{\mydata}, the first dataset for predicting the overall stance of full-length news articles in Korean. The dataset comprises 2,000 news articles, each manually annotated with its stance toward one of 47 nationwide issues. In addition to article-level annotations, the dataset provides stance labels for finer-grained components of the articles, including headlines, concluding paragraphs, and quotations within the body text. In total, \textsc{\mydata} contains 21,650 segment-level stance annotations.

Building on this dataset, we propose \textsc{\mymethod}, short for \textbf{Jo}urnalism-guided \textbf{A}gentic \textbf{I}n-\textbf{C}ontext \textbf{L}earning. As illustrated in Figure~\ref{fig:intro}, this LLM-based method performs article-level stance prediction via in-context learning while delegating stance classification of journalism-guided segments---such as leads and quotations---to a language model (LM) agent. The resulting segment-level predictions are then integrated with the primary LLM, enabling more accurate inference of an article's overall stance toward a target issue. Experimental results showed that \textsc{\mymethod} outperforms existing methods, including chain-of-thought~\cite{wei2022chain}, highlighting the effectiveness of segment-level agency in article-level stance detection.

We make the following key contributions.

\begin{itemize}
    \item We introduce \textsc{\mydata}, the first Korean dataset for article-level news stance detection, consisting of 2,000 articles and 21,650 segments annotated with stance labels.
    \item We propose \textsc{\mymethod}, an agentic in-context learning framework that predicts article-level stance by leveraging segment-level stance predictions generated by a language model agent.
    \item We present experimental results that demonstrate the effectiveness of segment-level agency in capturing the overall stance of long-form news articles, as well as its generalizability to another language. 
    \item We highlight the practical utility of \textsc{\mymethod} in fostering pluralistic and trustworthy media environments through two case studies.
\end{itemize}

\section{Related works}

\paragraph{Stance detection on news articles}
We review related studies on stance detection using news data. Much of this work has focused on news headlines~\cite{yoon2019detecting,bourgonje-etal-2017-clickbait,ghanem-etal-2018-stance,ferreira2016emergent}, reflecting the broader trend of applying stance detection to short texts such as tweets~\cite{mohammad2016dataset}. \citet{pomerleau2017fnc}, for example, introduced a dataset for classifying the stance of news headlines toward unverified claims. Other studies have extended this line of work to news sentences~\cite{mets2024automated,luusi-etal-2024-political} and full-length articles~\cite{conforti-etal-2020-stander,mascarell-etal-2021-stance}.

Closely related research has examined framing and media bias. \citet{card-etal-2015-media} introduced an annotated corpus of news articles covering 15 frames across diverse social issues, which has since become a key resource in automated frame analysis~\cite{kwak2020systematic, card-etal-2016-analyzing, roy-goldwasser-2020-weakly}. Other studies have sought to predict the political bias of news articles at different levels of granularity~\cite{baly-etal-2018-predicting,baly-etal-2020-detect,hong-etal-2023-disentangling,chen-etal-2020-analyzing}, with recent studies leveraging LLMs for bias prediction~\citet{lin-etal-2024-indivec}. 

In contrast to this body of research, our focus is on detecting the overall stance of entire news articles toward specific issues---a task that remains relatively understudied.

\paragraph{LLM-based stance detection}
Prior research on stance detection has primarily focused on modeling the relationship between a text and a target to infer the expressed stance~\cite{2020Kucuk,aldayal-2021-state}. Early approaches relied on bag-of-words representations~\cite{mohammad-etal-2016-semeval} and recurrent neural networks~\cite{augenstein-etal-2016-stance,du2017stance}. With the advent of pretrained language models, researchers began to explore masked language models (MLM)~\cite{he-etal-2022-infusing, chai-etal-2022-improving, li-caragea-2021-target,liu-etal-2022-politics}. For example, \citet{li-etal-2021-p} proposed an uncertainty-aware self-training method for BERTweet. Building on these advances, subsequent work has investigated instruction-tuned LLMs. \citet{zhu2023can} conducted a preliminary evaluation of in-context learning by ChatGPT, reporting performance that lagged behind human annotators, while \citet{cruickshank2023prompting} compared in-context learning and fine-tuning using open-weight LLMs. More recent studies have further advanced LLM-based approaches: \citet{lan-etal-2023-colla} proposed a multi-agent framework in which LLM experts collaborate on stance prediction; \citet{li-etal-2023-stance} incorporated retrieved and filtered background knowledge from Wikipedia; and \citet{zhang-etal-2024-llm-enabled} extracted diverse forms of stance-related knowledge from LLMs to train stance classifiers. Our study extends this line of research by introducing a novel stance detection method that leverages segment-level signals for article-level stance classification. 

\section{Problem and Dataset}

\subsection{Target Problem}

We address the task of stance detection in news articles, which involves identifying the positional stance of a news article toward a given social issue. Formally, given a news article $A$ covering a target issue $T$, the objective is to determine the overall stance of $A$ toward $T$. The stance label $L$ is categorized into one of three classes: \textit{supportive}, \textit{neutral}, or \textit{oppositional}. A stance detection model $f(\cdot)$, which takes $A$ and $T$ as inputs, is tasked with predicting $L$. Model performance is evaluated using standard classification metrics.

The target problem represents a specialized case of stance detection, an NLP task aimed at determining the position or attitude expressed in a text regarding a particular target~\cite{2020Kucuk,hardalov-etal-2022-survey,zhang2024survey}. While stance detection has been widely studied in the context of short-form content such as tweets, forum posts, or headlines, its application to long-form journalistic texts remains a formidable challenge due to the complex nature of news articles, which can be summarized in two key aspects. 

First, professional journalism typically privileges verification over assertion ~\cite{kovach2021elements}. Adhering to normative ideals of neutrality and balance, news articles often refrain from making overt evaluative claims. Instead, they rely on indirect cues, such as source selection ~\cite{zoch1998women, druckman2005impact}, narrative framing~\cite{nelson1997media, gentzkow2010drives}, and lexical subtleties~\cite{simon2007toward, schuldt2011global}, to communicate a stance, if any, toward a given issue. Even when an article expresses a positional preference, it is frequently nuanced, hedged, or ambivalent, making it difficult for models to detect without a deep understanding of rhetorical and discursive context.

Second and relatedly, the stance expressed in a news article is rarely concentrated in a single sentence or paragraph. Rather, it is often distributed across multiple textual layers, including headlines, leads, quotations, and framing devices. These layers may contain conflicting or ambiguous signals, especially in articles that attempt to present multiple sides of an issue. Accordingly, stance detection models must be capable of synthesizing fragmented and context-dependent cues across the entire document, a task made more challenging by the sheer length of the news texts.

Furthermore, despite recent advances in language understanding, LLMs often struggle to retain salient contextual information when processing long documents
~\cite{liu-etal-2024-lost}, leading to degraded performance~\cite{reuver-etal-2024-investigating}. This limitation is particularly pronounced in the news domain, where articles are significantly longer and more discursively layered than the short texts---such as tweets or single sentences---commonly used in prior stance detection research. 

To address these challenges, we propose a hierarchical modeling approach that first infers the stance at the level of smaller discourse units (e.g., paragraphs or sections), and subsequently integrates these local predictions to determine the overall stance of the article. This framework is designed to retain local context and capture dispersed stance cues in assessing how different parts of a news story contribute to its overall position on an issue.

\begin{table*}[t]
    \centering
    \small
        \begin{subtable}[b]{\linewidth}
    \begin{tabular}{p{15cm}}
         \makecell{\normalsize\textbf{Target Issue}: The National Assembly's Approval of the Ban on Dog Meat Consumption}\\\hline
         \textbf{Headline} (\emph{Supportive}): Will the History of `Dog Meat Consumption' end···Animal Rights Groups Welcome Passage of `Special Act Banning Dog Meat Consumption' \colorbox{cyan!30}{``Expect Practical Termination''}\\
         \textbf{Body Text}\\
         -\textsf{Lead} (\emph{Supportive}): Animal rights groups unanimously welcome National Assembly passage of `Dog Meat Ban Bill.' \colorbox{cyan!30}{``Practical effect of banning dog meat consumption, expect termination in the near future.''} Korea Dog Meat Association strongly opposes \colorbox{red!30}{``Depriving 10 million citizens of their right to food.''} The `Special Act on the Termination of Dog Breeding, Slaughter, Distribution, and Sale for Food Purposes' passed the National Assembly plenary session on the 9th. Animal rights groups that have been working to ban dog meat consumption welcomed the development, saying \colorbox{cyan!30}{``The passage of the special act will be the first step toward a country without dog meat consumption.''}\\
         -\textsf{Conclusion} (\emph{Neutral}): Previously, the Korea Dog Meat Association had announced in November last year that they would release 2 million dogs throughout Seoul in protest of the dog meat ban law near the Presidential Office in Yongsan, Seoul.\\
         \textbf{Overall Stance}: \emph{\textcolor{blue}{Supportive}}\\
         \hline
    \end{tabular}%
    \end{subtable}   
    \begin{subtable}[b]{\linewidth}
    \begin{tabular}{p{15cm}}
         \textbf{Headline} (\emph{Neutral}): Ban on Dog Meat Consumption: \colorbox{cyan!30}{``A Historic Victory''} vs \colorbox{red!30}{``Awaiting Constitutional Appeal''} \\
         \textbf{Body Text}\\
         -\textsf{Lead} (\emph{Neutral}): The passage of Korea's so-called `Dog Meat Ban Bill' on January 9 has triggered sharply divergent responses from advocacy groups and industry representatives. Animal rights organizations celebrated the vote as a watershed moment, declaring it \colorbox{cyan!30}{``A historic victory for animal rights.''} Meanwhile, the Korea Dog Meat Association, which has fiercely opposed the legislation, condemned the decision as an infringement on basic rights, stating that \colorbox{red!30}{``The freedom to choose one's occupation has been taken away.''} The group announced its intention to file a constitutional appeal.\\

         -\textsf{Conclusion} (\emph{Neutral}): The bill, formally titled, `The Special Act on the Termination of Dog Breeding, Slaughter, Distribution, and Sale for Food Purposes,' bans all commercial activities involving dogs for human consumption, including breeding, slaughter, distribution, and sale. \\
         \textbf{Overall Stance}: \textcolor{teal}{\emph{Neutral}}\\
         \hline
    \end{tabular}%
    \end{subtable}
    \begin{subtable}[b]{\linewidth}
    \begin{tabular}{p{15cm}}
         \textbf{Headline} (\emph{Oppositional}):  With the ban on Dog Meat Passed, \colorbox{red!30}{``Longtime Boshintang Vender Says, I'm at a loss''}\\
         \textbf{Body Text}\\
         -\textsf{Lead} (\emph{Oppositional}): Confusion and concern are growing among dog meat industry workers following the National Assembly's approval of a bill that will outlaw the dog meat consumption. Although enforcement measures and penalties won't take effect until 2027, the law has already sparked fresh controversy. While officials argue that the legislation will end decades of bitter debate, questions are mounting about how to manage the sudden increase in abandoned dogs and unregulated breeding facilities.\\
         -\textsf{Conclusion} (\emph{Oppositional}): {\raggedright Mr. B, a long-time vendor in the industry, said \colorbox{red!30}{``Once the suppliers disappeared, I had no} \colorbox{red!30}{ choice but to shut down for a while because I simply couldn't get any meat.''} He added, \colorbox{red!30}{``My rent is 1.6 million won }\\ \colorbox{red!30}{per month. To stay afloat, I need to make at least 800,000 won per day, but now I'm barely making 200,000. At this }\\\colorbox{red!30}{rate, I won’t be able to keep the doors open.''}}\\
         \textbf{Overall Stance}: \emph{\textcolor{red}{Oppositional}}\\
         \hline
    \end{tabular}%
    \end{subtable}
    \caption{Data examples translated in English, of which the remaining body text is omitted for brevity. The colored highlight indicates the stance label for quotations (blue: supportive, red: oppositional).}
    \vspace{-2mm}
    \label{tab:labeling_example}
\end{table*}

\subsection{Dataset: \textsc{\mydata}}

Although prior research has examined news stance detection~\cite{luusi-etal-2024-political, mascarell-etal-2021-stance, liu-etal-2022-politics}, most studies have focused on high-resource languages such as English and German. To address this limitation, we introduce a new annotated corpus in Korean. The dataset contains manually labeled stance annotations for sub-components of news articles, enabling fine-grained analysis and supporting the development of more advanced stance detection methods.

\paragraph{Raw data collection}

We collected Korean news articles published between June 2022 and June 2024 using BigKinds\footnote{https://www.bigkinds.or.kr/} and Naver News\footnote{https://news.naver.com/}. BigKinds, operated by the Korea Press Foundation (KPF)\footnote{https://www.kpf.or.kr/}, is a news platform that provides metadata (e.g., headlines, publishers) for weekly nationwide issues across diverse domains, including labor, gender, national, and international affairs. As a government-affiliated organization, the KPF curates a comprehensive news archive, ensuring that BigKinds captures social issues of national significance. From this archive, we randomly sampled 47 issues, maintaining temporal balance over the two-year period. Because BigKinds does not provide full text, we retrieved the corresponding content using the Naver News search API\footnote{https://developers.naver.com/}, a major news aggregator in Korea. The data collection comprises 2,989 articles from 31 news outlets, which were then manually annotated for stance in the subsequent step.

\paragraph{Manual annotation}

Following the lead of the third author---who holds a Ph.D. in mass communication and brings journalism expertise---we developed a manual annotation guideline for labeling both the overall stance of a news article toward its target issue and the stance expressed in its sub-components. The guideline is informed by narrative features known to signal stance, including information selection~\cite{nelson1997media,zoch1998women,gentzkow2010drives, druckman2005impact}, patterns of direct quotation~\cite{mcglone2005quoted, han2018polarizing, song2023detecting}, lexical choices~\cite{simon2007toward, schuldt2011global}, and cues implying preferred interpretations or intended actions~\cite{bolsen2015counteracting, mcintyre2019solutions}.

The annotation process consists of two main tasks. The first involves classifying each article into one of four categories based on established journalism genre distinctions: \emph{straight news} (factual reporting that objectively conveys facts without interpretation), \emph{analysis} (articles that extend beyond factual reporting to include context, implications, issue organization, or future scenarios), \emph{opinion} (articles predominantly expressing subjective opinions or interpretations), and \emph{other} (e.g., interviews). For stance annotation, we focus exclusively on \emph{analysis} and \emph{opinion} pieces, as these genres are more likely to contain opinionated content~\cite{alhindi-etal-2020-fact}. 

The second task involves assessing the article's stance toward a given issue, classifying it as \emph{supportive}, \emph{neutral}, or \emph{oppositional}. Each article in our dataset addresses one predefined issue, as identified by BigKinds.\footnote{Most articles in Korean journalism focus on a single issue~\cite{park2013dignity}.} In addition to article-level stance labels, we annotated four key structural components of each article: the headline, lead, conclusion, and direct quotations. When stance is ambiguous, annotators were encouraged to consult additional articles on the same issue to improve contextual understanding and ensure consistency in labeling.

Two annotators from our institution were recruited and trained to follow the annotation guidelines meticulously. The annotators labeled all 2,000 articles and 21,650 segments, achieving substantial inter-coder reliability, with Krippendorff's alpha ranging from 0.68 to 0.84 across different segments and article-level annotations. In cases of disagreement, annotation conflicts were resolved through discussion and consensus. The detailed guidelines and labeling interface are shown in Figure~\ref{fig: labeling_interface}. 

Table~\ref{tab:labeling_example} presents three annotation examples that illustrate different stances on the same issue. As shown in these examples, segment-level stance labels offer important cues for interpreting the overall position of a news article toward the issue. The original example in Korean is in Table~\ref{tab:labeling_example_korean_original}.

\paragraph{Dataset statistics}
The final dataset comprises 2,000 news articles covering 47 distinct issues. Following prior work on stance detection~\cite{reuver-etal-2024-investigating}, we divided the dataset into two splits such that each split contains a disjoint set of issues---24 for training and 23 for testing. Accordingly, the training and test sets consist of 999 and 1,001 articles, respectively. This issue-level split prevents models from relying on issue-specific cues when predicting stance labels. Table~\ref{tab:dataset_characteristics} presents descriptive statistics for the training and test sets, which are broadly comparable. On average, articles contain 1,483 characters, with lengths ranging from 376 to 8,185 characters. Each article includes an average of 7.6 direct quotations, with the number ranging from 0 to 39. Further analysis of label distributions and cross-segment associations is provided in Section~\ref{sec:app:dataset_details}.

\begin{table}[t]
    \centering
    \resizebox{0.99\linewidth}{!}{%
    \begin{tabular}{lccc}
    \hline
    \textbf{} & Train & Test & Total \\
    \hline
    \makecell[l]{\# Articles\\(S/N/O)} & \makecell{999\\(315/344/340)} & \makecell{1001\\(323/330/348)} & \makecell{2000\\(638/674/688)} \\
    \makecell[l]{\#  Issues} &  24 & 23 & 47 \\\hline
    \# Characters (max) & 5451 & 8185 & 8185 \\
    \# Characters (mean) & 1478 & 1489.14 & 1483.58 \\
    \# Characters (median) & 1348 & 1318 & 1335 \\
    \# Characters (min) & 376 & 413 & 376 \\\hline
    \# Quotations (max) & 32 & 39 & 39 \\
    \# Quotations (mean) & 7.63 & 7.54 & 7.59 \\
    \# Quotations (median) & 7 & 7 & 7 \\
    \# Quotations (min) & 0 & 0 & 0 \\
    \hline
    \end{tabular}
    }
\caption{Descriptive statistics of \textsc{\mydata} across data splits (S: Supportive, N: Neutral, O: Oppositional).}
    \label{tab:dataset_characteristics}
\end{table}

\section{Proposed Method: \textsc{\mymethod}}
LLMs can be adapted to new tasks without parameter updates by providing task instructions in the input prompt, a technique known as \emph{in-context learning}~\cite{brown2020language}. However, applying this approach to our target task is suboptimal due to the length and structural complexity of news articles, which often leads to context loss~\cite{liu-etal-2024-lost} and degraded performance~\cite{bertsch2024context}. To address this limitation, we propose \textsc{\mymethod} (\textbf{Jo}urnalism-guided \textbf{A}gentic \textbf{I}n-\textbf{C}ontext \textbf{L}earning), an agentic in-context learning framework that enhances LLM prompts by incorporating stance labels for shorter, journalism-guided structural segments of news articles. We describe this framework \emph{agentic} because the primary LLM for article-level stance detection delegates segment-level stance prediction to a separate language model (LM) agent.

\paragraph{Segment-level stance detection}
We employed an LM agent to infer stance labels for shorter structural segments of a given news article. We assumed that the agent can reliably predict the stance of these segments, thereby assisting the primary LLM in inferring the article's overall stance by leveraging these localized stance signals.

\begin{table*}[ht]
\centering
\begin{subtable}[t]{0.9\linewidth}
    \centering
    \resizebox{\textwidth}{!}{
    \begin{tabular}{cc|cc|cc|cc}
    \hline
    \multicolumn{2}{c|}{RoBERTa} & \multicolumn{2}{c|}{CoT Embeddings} & \multicolumn{2}{c|}{LKI-BART} & \multicolumn{2}{c}{PT-HCL} \\ \cline{0-7}
    Accuracy & F1 & Accuracy & F1 & Accuracy & F1 & Accuracy & F1 \\\hline
    0.594$\pm$0.029 & 0.577$\pm$0.046 & 0.582$\pm$0.028 & 0.562$\pm$0.039 & 0.545$\pm$0.011 & 0.538$\pm$0.014 & \textbf{0.617$\pm$0.007} & \textbf{0.618$\pm$0.008} \\\hline
    \end{tabular}
    }
    \caption{Existing methods}
    \label{tab:perf:existing}
    \vspace{2mm}
\end{subtable}
\hfill
\begin{subtable}[t]{0.99\textwidth}
    \centering
    \resizebox{\linewidth}{!}{
    \begin{tabular}{c|cc|cc|cc|cc|cc}
    \hline
    \multirow{2}{*}{Method} & \multirow{2}{*}{CoT} & \multirow{2}{*}{Few-shot} & \multicolumn{2}{c|}{GPT-4o-mini} & \multicolumn{2}{c|}{Gemini-2.0-flash} & \multicolumn{2}{c|}{Claude-3-haiku} & \multicolumn{2}{c}{EXAONE-2.4b (fine-tuned)} \\ \cline{4-11}
    & & & Accuracy & F1 & Accuracy & F1 & Accuracy & F1 & Accuracy & F1 \\ \hline
    \multirow{4}{*}{Baseline}
    &  &  &  0.594$\pm$0.002&  0.577$\pm$0.002
&  0.636$\pm$0.001&  0.628$\pm$0.002
& \textbf{0.568$\pm$0.002} & \textbf{0.538$\pm$0.002}
& \textbf{0.554$\pm$0.003} & \textbf{0.544$\pm$0.003}
\\ 
    & √ &  &  \textbf{0.597$\pm$0.002}&  \textbf{0.581$\pm$0.003}
&  \textbf{0.661$\pm$0.002} & \textbf{0.657$\pm$0.002}
&  0.561$\pm$0.002&  0.533$\pm$0.003
&  0.54$\pm$0.001&  0.539$\pm$0.003
\\
    &  & √ &  0.565$\pm$0.003&  0.54$\pm$0.003
&  0.635$\pm$0.003&  0.631$\pm$0.004
&  0.545$\pm$0.004&  0.523$\pm$0.005
&  0.445$\pm$0.004&  0.431$\pm$0.003
\\
    & √ & √ &  0.526$\pm$0.003&  0.494$\pm$0.003&  0.641$\pm$0.005&  0.635$\pm$0.004&  0.541$\pm$0.004&  0.51$\pm$0.004&  0.456$\pm$0.002&  0.443$\pm$0.002\\
    \hline
    \multirow{4}{*}{\makecell{\textsc{\mymethod}\\(Oracle)}}
    &  &  &  0.724$\pm$0.001&  0.706$\pm$0.001
&  0.758$\pm$0.001&  0.753$\pm$0.001
&  0.791$\pm$0.002&  0.789$\pm$0.002
&  \textbf{0.837$\pm$0.004}&  \textbf{0.837$\pm$0.004}
\\ 
    & √ &  &  0.716$\pm$0.002&  0.692$\pm$0.002
&  \textbf{0.778$\pm$0.001} & \textbf{0.776$\pm$0.001}
&  0.74$\pm$0.001&  0.732$\pm$0.001
&  0.813$\pm$0.003&  0.821$\pm$0.002
\\
    &  & √ &  \textbf{0.796$\pm$0.003} & \textbf{0.798$\pm$0.003} &  0.772$\pm$0.003&  0.769$\pm$0.003
&  \textbf{0.815$\pm$0.001} &  \textbf{0.816$\pm$0.001}
&  0.344$\pm$0.004&  0.308$\pm$0.004
\\
    & √ & √ &  0.747$\pm$0.001&  0.74$\pm$0.002&  0.777$\pm$0.003&  0.773$\pm$0.003&  0.794$\pm$0.002&  0.797$\pm$0.002&  0.338$\pm$0.005&  0.3$\pm$0.006\\
    \hline
    \multirow{4}{*}{\makecell{\textsc{\mymethod}\\(RoBERTa)}}
    &  &  &  0.571$\pm$0.002&  0.53$\pm$0.001
&  0.633$\pm$0.003&  0.619$\pm$0.004
&  0.591$\pm$0.004&  0.577$\pm$0.004
&  0.584$\pm$0.002&  0.581$\pm$0.001
\\ 
    & √ &  &  0.553$\pm$0.001&  0.509$\pm$0.001
&  0.633$\pm$0.001&  0.626$\pm$0.001
&  0.579$\pm$0.003&  0.552$\pm$0.004
&  \textbf{0.601$\pm$0.006} & \textbf{0.599$\pm$0.006}
\\
    &  & √ &  \textbf{0.607$\pm$0.006} & \textbf{0.602$\pm$0.007}
&  0.662$\pm$0.001&  0.657$\pm$0.001
&  \textbf{0.639$\pm$0.002}&  \textbf{0.638$\pm$0.004}
&  0.354$\pm$0.003&  0.331$\pm$0.003
\\
    & √ & √ &  0.591$\pm$0.002&  0.57$\pm$0.002 
    &  \textbf{0.678$\pm$0.002}&  \textbf{0.672$\pm$0.002}
    &  0.61$\pm$0.004&  0.608$\pm$0.004
&  0.332$\pm$0.001&  0.308$\pm$0.001\\
    \hline
    \end{tabular}
    }
    \caption{LLM in-context learning methods}
    \label{tab:perf:llm}
\end{subtable}
\caption{Performance for predicting overall stance of news articles measured on the test split of \textsc{\mydata}. The best model for each configuration is highlighted as bold.}
\vspace{-2mm}
\label{tab:overall_stance_perf}
\end{table*}

Specifically, we analyzed the following sub-components of news articles, grounded in journalism research~\cite{mencher1997news, bbc-paraphrasing}:
\begin{itemize}
    \item \textsf{Headline}: The title of the article, which conveys the core message and is designed to be clear and easily understood at a glance.
    \item \textsf{Lead}: The first paragraph of the article, which follows the inverted pyramid structure by summarizing the most important information and typically addressing at least three of the six classic questions (5Ws and 1H): Who, What, Where, When, Why, and How. 
    \item \textsf{Conclusion}: The final paragraph of the article, which often reinforces the main points or offers closing context or interpretation.
    \item \textsf{Quotations}: Direct speech from sources, enclosed in double quotation marks. Journalistic accounts are constructed largely through what sources say, rather than through direct observation or objective reality~\cite{sigal1986sources}. Source selection thus plays a critical role in shaping the stance and framing of news coverage~\cite{entman2004projections, schudson2003sociology}. Direct quotations are not only among the most commonly used evidentiary devices in news stories~\cite{tuchman-1978-making}, but also play a prominent role in shaping news narratives~\cite{song2023detecting, missouri-2013-news}.
\end{itemize}

We considered two types of LMs for segment-level stance prediction: (1) an LLM that performs in-context learning without parameter updates, and (2) a fine-tuned MLM trained on the segment-level annotations. According to the comparison experiments, we adopted a fine-tuned MLM for the best-performing variant.

\paragraph{Article stance prediction} To predict the overall stance of an article toward a target issue, we prompted an LLM with task instructions while augmenting segment-level stance labels predicted by an LM agent. These labels were embedded into the article using an XML-like format, enabling the LLM to incorporate them as contextual cues during inference. The proposed method is model-agnostic and can be applied to any instruction-following LLM. The prompt format and an example are shown in Figure~\ref{fig:proposed_prompt}.

\section{Evaluation}

We conducted evaluation experiments to assess the effectiveness of the proposed method for article-level news stance detection in comparison to existing approaches. Detailed experimental settings are provided in Section~\ref{sec:app:exp_setting}.

\subsection{Baseline Methods}

We employed four fine-tuned methods as baselines, each demonstrating state-of-the-art performance on stance detection benchmarks: \textbf{(1)}~\textsf{RoBERTa}~\cite{liu2019roberta} is a fine-tuned MLM for article-level stance detection. \textbf{(2)}~\textsf{CoT Embeddings}~\cite{gatto-etal-2023-chain} is a fine-tuned RoBERTa model on the explanation trace of an LLM for determining the stance of a given news article. \textbf{(3)}~\textsf{LKI-BART}~\cite{zhang-etal-2024-llm-driven} is an encoder-decoder model that incorporates contextual knowledge from an LLM into stance detection by prompting the LLM with both the input and target. \textbf{(4)}~\textsf{PT-HCL}~\cite{10.1145/3485447.3511994} is a hierarchical contrastive learning method designed to distinguish between target-invariant and target-specific features. Model checkpoints are provided in Section~\ref{sec:app:exp_setting}.

\subsection{Results}

Table~\ref{tab:overall_stance_perf} presents the evaluation results of the baseline and proposed methods on the test set of \textsc{\mydata}. Table~\ref{tab:perf:existing} reports the performance of four existing state-of-the-art methods, trained on the training split. Table~\ref{tab:perf:llm} summarizes the performance of LLM-based in-context learning methods, including an instruction-only baseline and two variants of the proposed \textsc{\mymethod}, listed in separate rows. The second and third columns indicate whether each model employs advanced prompting techniques---chain-of-thought reasoning~\cite{wei2022chain} (CoT) or few-shot sample augmentations ($k$ = 6)~\cite{brown2020language}. The remaining columns present results across four LLM backbones: three proprietary models---GPT-4o-mini~\cite{hurst2024gpt}, Gemini-2.0-flash~\cite{team2023gemini}, and Claude-3-haiku~\cite{anthropic2024claude}---used without parameter updates, and one open-weight model, EXAONE-2.4b~\cite{research2024exaone}, which was instruction fine-tuned. These models were selected for their strong performance in Korean language understanding. 

We present the performance of two variants of \textsc{\mymethod}, which differ in the source of segment-level stance predictions. The first assumes an idealized \textit{oracle} setting, where segment-level stance labels are perfectly accurate, providing an upper bound on model performance. The second replaces the oracle with a fine-tuned RoBERTa model trained on segment-level annotations from the training split of \textsc{\mydata}, representing a more realistic prediction scenario. RoBERTa was selected as the representative segment-level agent due to its competitive performance compared to the LLM-based agents, as shown in Table~\ref{tab:app:overall_stance_perf_lmm}.

Three key observations emerged from Table~\ref{tab:overall_stance_perf}. First, among the baseline methods, PT-HCL achieved the highest performance, with an accuracy of 0.617 and an F1 score of 0.618, followed by the fine-tuned RoBERTa model. These results highlight the effectiveness of contrastive learning and standard MLM fine-tuning for stance detection. 

Second, among the LLM-based in-context learning baselines, Gemini-2.0-flash stood out as the best-performing LLM when combined with CoT prompting, achieving an accuracy of 0.661 and F1 of 0.657. This outperformed all fine-tuned baselines. However, the effectiveness of CoT and few-shot prompting varied across LLMs, indicating model-specific sensitivity to prompting strategies. 

Third, incorporating the RoBERTa-based segment-level agent in \textsc{\mymethod} consistently improved stance detection performance across all LLMs; it yielded gains of up to +0.071 in accuracy and +0.1 in F1 compared to the in-context learning baselines. In this configuration, Gemini-2.0-flash again achieved the best performance when paired with CoT prompting and six-shot augmentation. However, the overall performance remains constrained by the quality of segment-level predictions, as reflected in the persistent performance gaps between this setting and the oracle variant across all LLMs. The largest gap of +0.236 in accuracy and +0.238 in F1 was observed for the fine-tuned EXAONE-2.4b, which achieved the highest performance in the oracle setting despite its relatively small model size. This result highlights the potential of \textsc{\mymethod} for efficient deployment in resource-constrained scenarios.

\begin{table}[t]
\centering
\begin{subtable}[t]{0.9\linewidth}
    \centering
    \resizebox{.9\linewidth}{!}{
    \begin{tabular}{lcc}
    \hline
    \makecell[c]{Model} & Accuracy & F1 \\\hline
    \makecell[l]{\textsc{\mymethod}} & 0.837$\pm$0.004&  0.837$\pm$0.004
\\
    w/o Headline& 0.827$\pm$0.002& 0.827$\pm$0.001\\
    w/o Lead & 0.763$\pm$0.001& 0.75$\pm$0.001\\
    w/o Conclusion & 0.767$\pm$0.003& 0.77$\pm$0.003\\
    w/o Quotations & 0.828$\pm$0.004& 0.826$\pm$0.003\\\hline
    \end{tabular}
    }
    \caption{\textsc{\mymethod} (Oracle)}
    \vspace{1mm}
\end{subtable}
\hfill
\begin{subtable}[t]{0.9\linewidth}
    \centering
    \resizebox{.9\linewidth}{!}{
\begin{tabular}{lcc}
    \hline
    \makecell[c]{Model} & Accuracy & F1 \\\hline
    \makecell[l]{\textsc{\mymethod}} & 0.678$\pm$0.002&  0.672$\pm$0.002
\\
    w/o Headline& 0.672$\pm$0.003& 0.67$\pm$0.004\\
    w/o Lead & 0.61$\pm$0.001& 0.63$\pm$0.002\\
    w/o Conclusion & 0.663$\pm$0.003& 0.658$\pm$0.005\\
    w/o Quotations & 0.676$\pm$0.002& 0.669$\pm$0.002\\\hline
    \end{tabular}
    }
    \caption{\textsc{\mymethod} (RoBERTa)}
\end{subtable}
\caption{Impact of segment label ablation on stance detection performance.}
\vspace{-2mm}
\label{tab:abl}
\end{table}

\paragraph{Ablation study on segment labels}
We conducted an ablation study to examine the contribution of stance labels from individual news segments to overall article-level stance prediction. Specifically, we evaluated model variants in which the stance label of a particular segment was omitted from the prompt. Table~\ref{tab:abl} reports results for two settings: the oracle variant of \textsc{\mymethod} using EXAONE-2.4b, and the best-performing RoBERTa-agent variant, which employs Gemini-2.0-flash as the backbone LLM with CoT prompting and six-shot sample augmentation. In each table, the second row presents the performance of \textsc{\mymethod} in its best configuration, followed by ablated variants. In the oracle setting, removing the lead segment resulted in the largest performance drop, while removing the headline or quotations yielded the smallest. When using predicted segment-level labels, the performance gaps were slightly reduced, but the decreasing trend due to ablation persisted. In this setting, removing the lead segment again caused the largest drop, whereas removing direct quotations resulted in the smallest, suggesting that quote-level stances are harder to interpret due to their brevity and subtlety. These findings highlight the importance of incorporating multiple segments in context, rather than focusing on a single segment---such as news headlines, as commonly addressed in prior stance detection research~\cite{ferreira2016emergent,bourgonje-etal-2017-clickbait}. Additional results are provided in Section~\ref{sec:app:results}, where we further confirm the effectiveness of journalism-guided segments compared to random segment-based label augmentation.

\begin{figure}[t]
     \centering
     \begin{subfigure}[b]{0.49\linewidth}
         \centering
         \includegraphics[width=.99\linewidth]{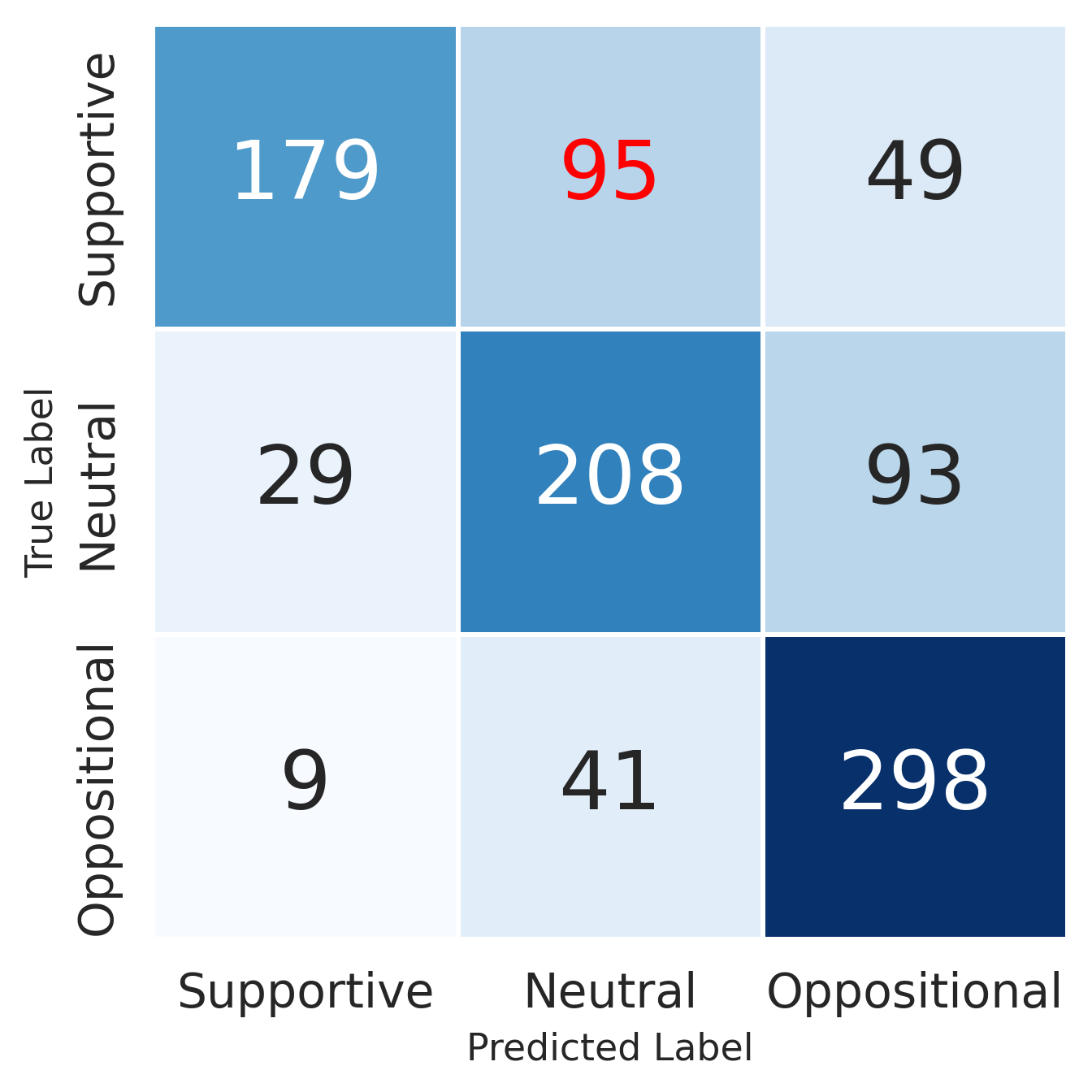}
        \caption{\textsc{\mymethod}}
     \end{subfigure}
     \begin{subfigure}[b]{0.49\linewidth}
         \centering
         \includegraphics[width=.99\linewidth]{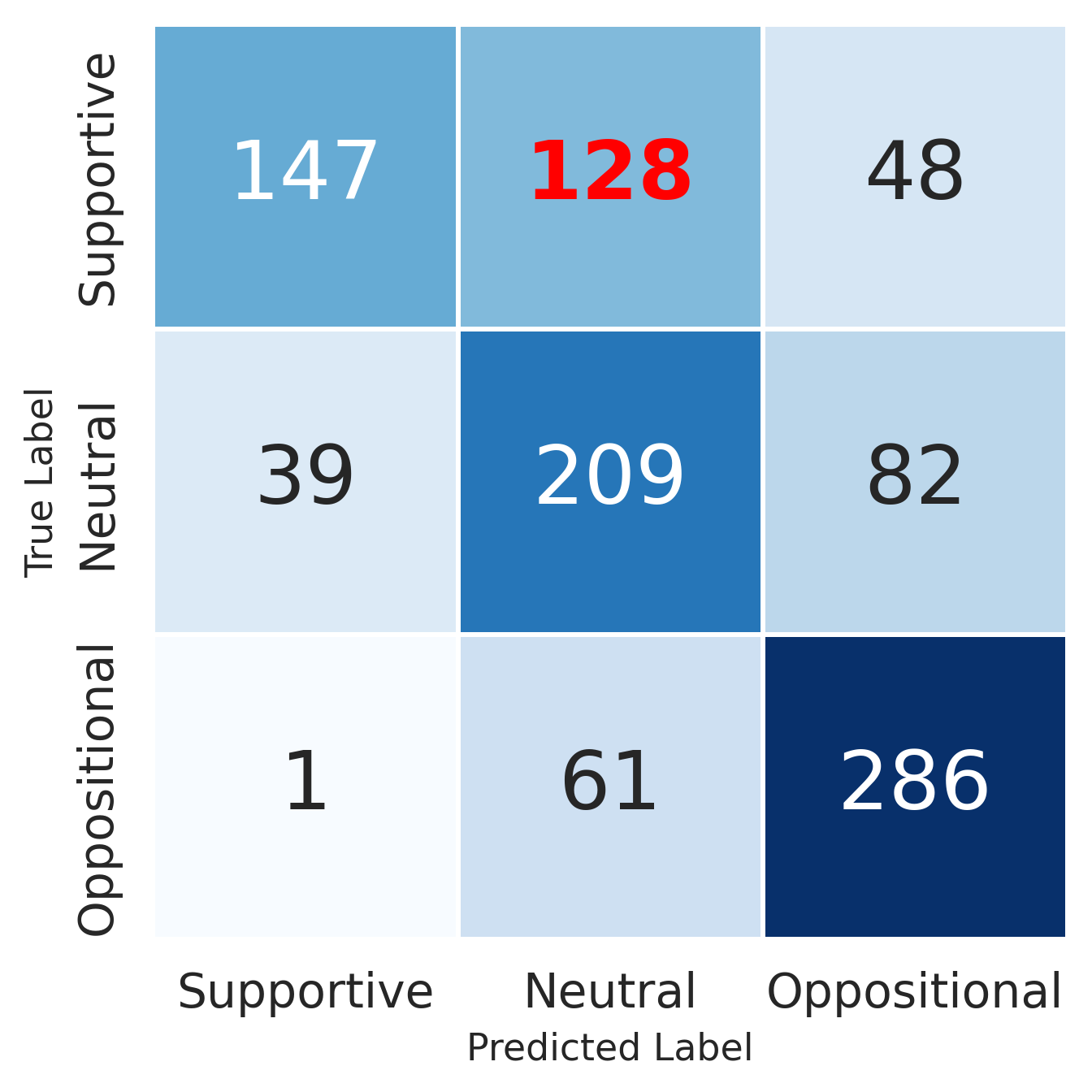}
        \caption{Baseline}
     \end{subfigure}
\caption{Confusion matrices of the baseline and proposed methods, illustrating the challenges in identifying cues indicative of supportive stances.}
\vspace{-2mm}
\label{fig:conf_matrix}
\end{figure}

\paragraph{Error analysis}
Figure~\ref{fig:conf_matrix} presents the confusion matrices for the best-performing variant of \textsc{\mymethod} using the RoBERTa agent and a baseline in-context learning method, both employing Gemini-2.0-flash as backbone. The results showed that the proposed method achieves higher accuracy across all three target classes. However, both models exhibited the greatest difficulty in correctly classifying articles labeled as \emph{supportive}, followed by \emph{neutral} and \emph{oppositional}. Among the 323 articles labeled as supportive, the baseline method frequently misclassified them as \emph{neutral} (128 instances) or \emph{oppositional} (48 instances). While \textsc{\mymethod} performed better overall, it exhibited a similar error pattern, highlighting the challenge of identifying cues indicative of supportive stances---an area warranting further investigation. We provide qualitative error analyses of representative misclassification cases in Section~\ref{sec:app:results}.

\paragraph{Generalization to another language}
To assess the generalizability of \textsc{\mymethod} beyond Korean and our dataset, we conducted an additional experiment on CheeSE, a German-language dataset for article-level news stance detection~\cite{mascarell-etal-2021-stance}. Since CheeSE does not provide segment-level annotations like \textsc{\mydata}, we adopted a distant supervision approach, training a RoBERTa model as the segment-level agent by assigning article-level stance labels to all segments. As shown in Table~\ref{tab:app:generalization_saas}, \textsc{\mymethod} outperformed both fine-tuned models and zero-shot prompting by a substantial margin. Among the three LLM backbones, Gemini-2.0-flash stood out as the best model, consistent with our findings on \textsc{\mydata}.

\begin{table}[t]
\centering
\begin{subtable}[t]{0.88\linewidth}
\centering
\resizebox{\linewidth}{!}{
\begin{tabular}{ccc}
\hline
Method & Accuracy & F1 \\\hline
RoBERTa &  \textbf{0.526$\pm$0.014} & \textbf{0.442$\pm$0.034}\\
CoT Embeddings &  0.424$\pm$0.01 & 0.198$\pm$0.003\\
LKI-BART & 0.466$\pm$0.01 & 0.34$\pm$0.011\\
PT-HCL & 0.517$\pm$0.021 & 0.39$\pm$0.048\\\hline
\end{tabular}
}
\caption{Fine-tuned models}
\label{tab:finetuned on german}
\end{subtable}

\vspace{2mm} 

\begin{subtable}[t]{0.99\linewidth}
\centering
\resizebox{\linewidth}{!}{
\begin{tabular}{cccc}
\hline
LLM & Method & Accuracy & F1 \\\hline
\multirow{2}{*}{GPT-4o-mini} & Zero-shot & 0.549$\pm$0.002 & 0.518$\pm$0.002 \\
& \textsc{\mymethod} & 0.593$\pm$0.002 & 0.538$\pm$0.002 \\\hline
\multirow{2}{*}{Gemini-2.0-flash} & Zero-shot & 0.566$\pm$0.001 & 0.54$\pm$0.001 \\
& \textsc{\mymethod} & \textbf{0.614$\pm$0.001} & \textbf{0.579$\pm$0.001} \\\hline
\multirow{2}{*}{Claude-3-haiku} & Zero-shot & 0.487$\pm$0.002 & 0.444$\pm$0.002 \\
& \textsc{\mymethod} & 0.59$\pm$0.001 & 0.527$\pm$0.001  \\\hline
\end{tabular}
}
\caption{In-context learning}
\label{tab:inference on german}
\end{subtable}
\caption{Stance detection performance on CheeSE, a German-language dataset, demonstrating the generalizability of \textsc{\mymethod} to a different language.}
\vspace{-2mm}
\label{tab:app:generalization_saas}
\end{table}

\section{Case studies}

We conducted two case studies to highlight potential applications of \textsc{\mymethod}. We additionally collected recent news data for six randomly selected issues from July 2024 to April 2025. The stance labels were manually annotated by the same annotators involved in the primary dataset.

\begin{table}[t] 
\centering
\resizebox{\linewidth}{!}{
\begin{tabular}{l|cc|cc}
\hline
    \multirow{2}{*}{\makecell{Method}} & \multicolumn{2}{c|}{$k=$5} & \multicolumn{2}{c}{$k=$10}\\\cline{2-5}
    & Diversity & Precision & Diversity & Precision \\\hline
    Contriever & 0.535 & \textbf{1} & 0.723 & \textbf{0.983} \\ \hline
    $+$ MMR & 0.622 & 0.975 & 0.764 & 0.969 \\ 
    $+$ MMR (\mymethod) & \textbf{0.647} & 0.983 & \textbf{0.793} & 0.971 \\\hline
\end{tabular}
}
\caption{Simulated results on the impact of incorporating predicted stances into news recommendations, highlighting the potential to promote political diversity.}
\label{tab:case_study_1_result}
\end{table}

\paragraph{Diversity in news recommendation}
The first case study investigated whether stance predictions by \textsc{\mymethod} could enhance political diversity in news recommendations. We assumed a scenario in which ten different users were each recommended a set of news articles after reading an initial article. As a baseline recommender based on content similarity, we used a multilingual version of Contriever~\cite{izacard2022unsupervised} to retrieve the top-20 most similar articles for each user from a newly collected article pool covering four distinct issues. 

We then applied two versions of the Maximum Marginal Relevance (MMR) re-ranking method~\cite{carbonell1998use} to these initial recommendations, using SentenceBERT as the embedding model~\cite{reimers-gurevych-2019-sentence}. The first was the standard MMR approach, which ranked articles based on embedding similarity. The second, denoted as MMR (\mymethod), incorporated predicted stance labels from \textsc{\mymethod}, encoded as one-hot vectors, to promote stance diversity during re-ranking. 

Table~\ref{tab:case_study_1_result} presents the evaluation results, reporting the average values of Diversity and Precision@$K$ for varying values of $K$ (from 5 to 10). Diversity was measured as the entropy of the political preference distribution among the recommended articles, with higher entropy indicating greater ideological diversity. Since the political leaning associated with each stance label can vary by issue, we manually mapped each stance label to one of three political categories: progressive, moderate, and conservative. 

The results indicated that re-ranking with stance predictions from \textsc{\mymethod} led to higher diversity scores, with only a slight reduction in precision compared to the baseline. Furthermore, the proposed re-ranking approach achieved a comparable level of precision to the standard MMR while yielding higher diversity. These findings demonstrate the potential of \textsc{\mymethod} for promoting politically diverse news recommendations.

\begin{figure}[t]
    \centering
    \includegraphics[width=\linewidth]{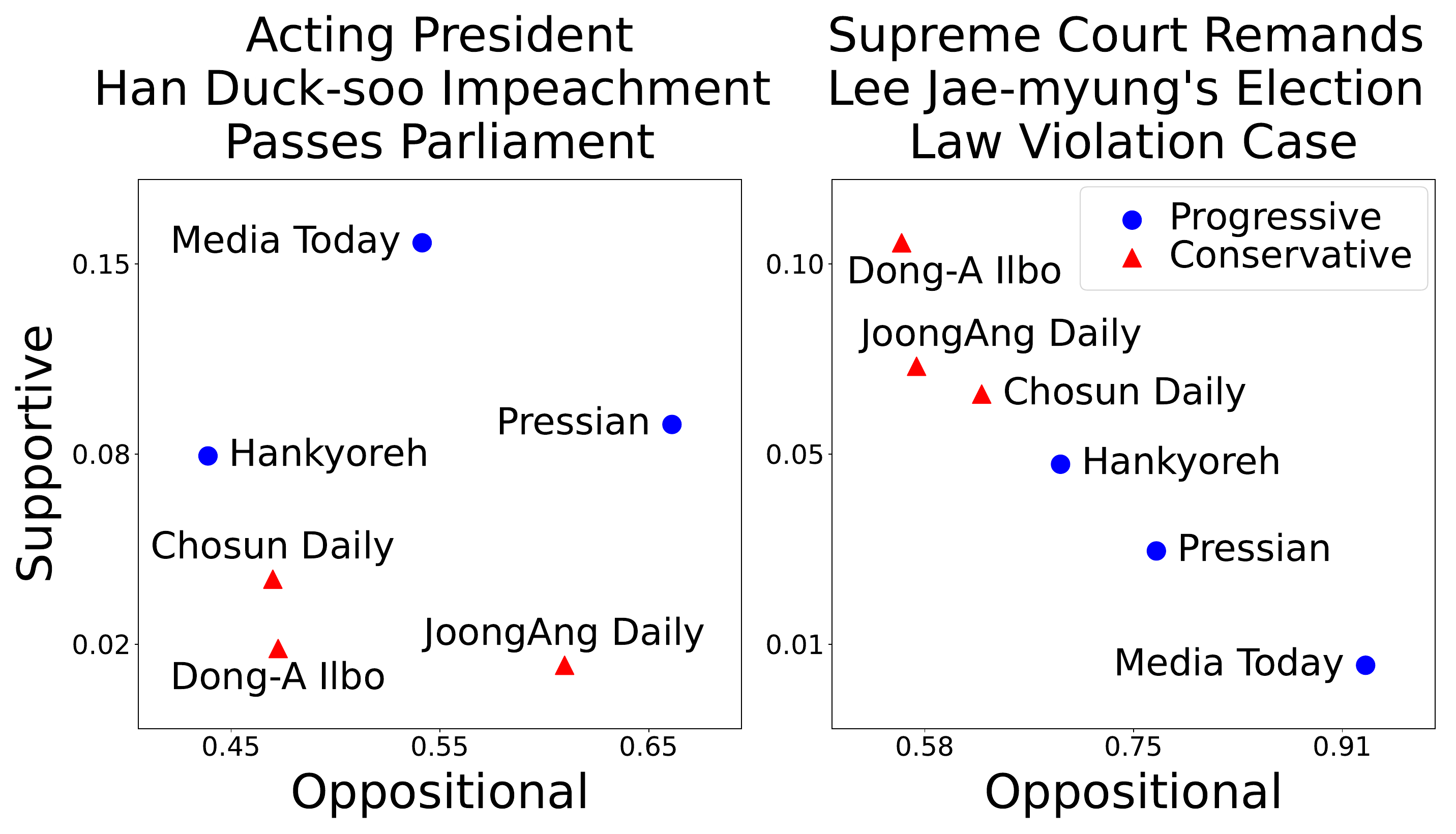}
\caption{Distribution of predicted stance labels for two recent issues not covered by \textsc{\mydata}, grouped by the political leaning of six major news outlets. This demonstrates the potential of \textsc{\mymethod} as an analytical tool for examining media bias.}
    \label{fig:case_study_2_result} 
\end{figure}

\paragraph{Political bias in news outlets}
The second case study demonstrated the utility of \textsc{\mymethod} as an analytical tool for identifying media bias. Figure~\ref{fig:case_study_2_result} presents a scatterplot in which each point reflects the distribution of predicted \textit{supportive} and \textit{oppositional} stances across news articles published by six major Korean news outlets. The analysis centered on two salient social issues tied to the 2025 presidential election. Outlets were categorized as either progressive or conservative based on ideological classifications established in prior literature~\cite{han2023news,song2007internet,JoDongSi2003HyeonYiSeop}. The resulting clusters revealed clear differences in stance patterns that align with each outlet's known editorial stance. These findings highlight the potential of \textsc{\mymethod} to map partisan bias in news coverage and support large-scale analyses of the media bias landscape.

\section{Conclusion}
This paper presented \textsc{\mydata}, a novel dataset for news stance detection in Korean, featuring stance annotations for both whole articles and journalism-guided news segments. Building on this resource, we proposed an agentic in-context learning method that enhances article-level stance detection by LLMs through the augmentation of segment-level stance labels generated by an LM agent. 
Experimental results demonstrated the effectiveness of incorporating journalism-guided segments and agentic in-context learning for stance prediction, as well as the method's generalizability across the dataset and language. Additionally, two case studies illustrated the broader applicability of the proposed dataset and method beyond benchmarking, supporting efforts toward fostering a pluralistic and credible media environment.

\section*{Limitations and Future Directions}
Our primary evaluation relies on \textsc{\mydata}, reflecting the fact that this is the first resource to pair journalism-guided segment-level labels with article-level stances. While the consistent performance gains observed across five different LLMs and an additional experiment involving a German-language dataset using distant supervision suggests that the core ideas generalize beyond this specific corpus, further validation across additional languages and media ecosystems remains essential. To support such research, we publicly release our annotation protocol, translated into English.

Our approach currently incurs additional inference costs due to the invocation of an LM agent at the segment level. Encouragingly, under the oracle setting, EXAONE-2.4b achieves the highest performance, indicating the potential for lightweight deployment. Moreover, techniques such as post-training quantization~\cite{lin2024awq} may help reduce inference overhead without compromising accuracy. A detailed comparison of the trade-offs between accuracy and computational cost is provided in Table~\ref{tab:app:comparison_tradeoff_perf_inf-cost}.

The two case studies serve as proof of concept rather than exhaustive evaluations. They demonstrate the method's potential for supporting politically diverse article recommendations and large-scale analyses of media bias, but do not yet encompass the full spectrum of real-world news genres and delivery platforms. Future work could extend these studies to more diverse scenarios to better assess the method's broader impact.

Finally, the proposed agentic in-context learning method could be extended to LLM-based multi-agent systems---for instance, through a debate-style framework~\cite{lan2024stance} in which each agent is assigned a distinct role grounded in journalistic principles.

\section*{Ethics Statement}

We constructed \textsc{\mydata} to support the training and evaluating article-level stance detection models, based on publicly available news articles retrieved via API. Since these articles are produced under strict journalistic standards, the use of this data raises minimal privacy concerns. While the primary purpose of the dataset is to support article-level stance detection, it is also suitable for segment-level stance detection, as demonstrated in our training of a fine-tuned RoBERTa model for segment-level prediction. Beyond benchmarking, \textsc{\mydata} has broader applicability for developing and evaluating stance detection models that contribute to pluralistic and credible media environments, as illustrated in our two case studies. The dataset will be released exclusively for academic purposes---such as benchmarking and media research---to respect the intellectual property rights of the original news publishers. Two graduate students (one female and one male) in an author's institution were recruited for manual annotation. In compliance with local wage regulations, they were compensated at a rate of approximately USD 7 per hour. Language editing was conducted using ChatGPT. This study was approved by the Institutional Review Board of Soongsil University.

\section*{Acknowledgments}
Dahyun Lee and Jonghyeon Choi contributed equally to this work. This research was supported by the MSIT(Ministry of Science and ICT), Korea, under the Graduate School of Metaverse Convergence support (IITP-2025-RS-2024-00430997) and Innovative Human Resource Development for Local Intellectualization (IITP-2025-RS-2022-00156360) programs, supervised by the IITP(Institute for Information \& Communications Technology Planning \& Evaluation). It was also supported by a grant from the National Research Foundation of Korea (NRF), funded by the Korea government (MSIT) (RS-2023-00252535). Kunwoo Park and Jiyoung Han are the corresponding authors.

\bibliography{custom}

\appendix

\counterwithin{figure}{section}
\counterwithin{table}{section}
\renewcommand\thefigure{A\arabic{figure}} 
\renewcommand\thetable{A\arabic{table}}

\section{Appendix}

\subsection{Experimental Setups}
\label{sec:app:exp_setting}

For evaluation, we used macro F1 and accuracy, which are standard metrics for multi-class classification. We reported the average performance over ten runs, along with the standard error, by varying the random seed from 42 to 51. The training split was used to train the fine-tuned baseline models and the segment-level stance prediction agents. Few-shot samples were selected from the training set via similarity search, using KLUE-RoBERTa-large as the dense retriever.

For the experiments on CheeSE, we used 1,762 samples by excluding 503 \emph{unklar} (unclear) and 1,428 \emph{Kein Bezug} (unrelated) samples. To train the LM agent, we split the data into 800/200/762 for the train, validation, and test sets, respectively, while preserving the label distributions across the splits.

Experiments were conducted using a machine equipped with three Nvidia RTX A6000 GPUs (48GB each) and 128GB of RAM. All experiments were run in a software environment configured with Python 3.9.19, PyTorch 2.5.1, Transformers 4.52.0, and vLLM 0.8.5. We accessed GPT-4o-mini, Claude-3-haiku, and Gemini-2.0-flash via API. We set the temperature as 1.0, and max tokens as 1000 for chain-of-thought prompting and 100 for others for all LLM API calls. For the full fine-tuning of EXAONE-3.5-2.4b, we used the AdamW optimizer with a learning rate of 5e-5, weight decay of 0.01, and 100 warmup steps. Training was conducted for 10 epochs with a per-device batch size of 6 for both training and evaluation.

For the RoBERTa-based models involving \textsc{\mydata}---including the segment-level stance agent and three fine-tuned baselines (RoBERTa, CoT Embeddings, PT-HCL)---we used KLUE-RoBERTa-large, a pretrained checkpoint trained on a Korean corpus~\cite{park2klue}. We employed XLM-RoBERTa-Large for the experiments on CheeSE. Based on validation experiments, we set the learning rate as $3\times10^{-5}$, with a batch size of 32 for CoT Embeddings and 16 for all other models. AdamW was used for optimizer and froze the bottom seven layers. GPT-4o-mini was used for CoT Embeddings and LKI-BART. We employed KoBART-base-v2 and BART-qg-German for LKI-BART. For training KoBART, we set the learning rate as $3\times10^{-5}$, batch size of 16, and used AdamW optimizer. We used mContriever and KR-SBERT-V40K-klueNLI-augSTS for case studies. We used $\lambda=0.3$ as the diversity hyperparameter in MMR. The hyperparameters were selected based on the settings reported in the original studies that introduced these methods.

The model ids and parameter sizes used in the experiments are provided below.

\begin{itemize}
    \item GPT-4o-mini: \texttt{gpt-4o-mini-2024-07-18} (Parameter size: unknown)
    \item Claude-3-haiku: \texttt{claude-3-haiku-20240307} (Parameter size: unknown)
    \item Gemini-2.0-flash: \texttt{gemini-2.0-flash} (Parameter size: unknown)
    \item EXAONE-3.5-2.4b: \url{https://huggingface.co/LGAI-EXAONE/EXAONE-3.5-2.4B-Instruct} (Parameter size: 2.41B)
    \item KLUE-RoBERTa-large: \url{https://huggingface.co/klue/roberta-large} (Parameter size: 337M)
    \item KLUE-RoBERTa-base: \url{https://huggingface.co/klue/roberta-base} (Parameter size: 111M)
    \item XLM-RoBERTa-large: \url{https://huggingface.co/FacebookAI/xlm-roberta-large} (Parameter size: 561M)
    \item KoBART-base-v2: \url{https://huggingface.co/gogamza/kobart-base-v2} (Parameter size: 124M)
    \item BART-qg-German: \url{https://huggingface.co/su157/bart-qg-german} (Parameter size: 139M)
    \item mContriever: \url{https://huggingface.co/facebook/mcontriever} (Parameter size: 178M)
    \item KR-SBERT-V40K-klueNLI-augSTS: \url{https://huggingface.co/snunlp/KR-SBERT-V40K-klueNLI-augSTS} (Parameter size: 116M)
\end{itemize}

\subsection{Supplementary Results}
\label{sec:app:results}

\begin{table}[h]
\centering
\resizebox{\linewidth}{!}{
\begin{tabular}{lcc}\hline
\makecell[c]{Label augmentation} & Accuracy & F1 \\\hline
\makecell[c]{For journalism-guided segments} & 0.678$\pm$0.002 & 0.672$\pm$0.002 \\
\makecell[l]{For randomly selected segments} & 0.649$\pm$0.001 & 0.645$\pm$0.002 \\\hline
\end{tabular}
}
\caption{Performance when ablating journalism-guided segment labels. The second row reports the performance of \textsc{\mymethod}, while the third row shows the performance of a counterpart model where labels for randomly selected segments are augmented.}
\label{tab:abl-journalism}
\end{table}

\paragraph{Effects of journalism-guided segments}
Table~\ref{tab:abl-journalism} presents the results of an ablation experiment to understand the impact of journalism-guided segments integrated with \textsc{\mymethod}. The second row reports the best performance of the proposed method using Gemini-2.0-flash as the backbone LLM, with chain-of-thought prompting and six-shot sample augmentation. The third row presents the performance of a counterpart model in which randomly selected sentences were used as targets for label augmentation. To control for the length effects, sentences were sampled without replacement until the total length matches that of the journalism-guided segments. The results indicated a performance drop---0.029 in accuracy and 0.027 in F1---demonstrating the advantage of journalism-guided inductive bias in selecting target segments.

\paragraph{Varying performance by article types}
We examined how performance varied across different article types, focusing specifically on \emph{opinion} and \emph{analysis} articles. Previous research has suggested that the two genres exhibit significant differences in how leads are used, one of the journalism-guided segments considered for \textsc{\mymethod}. While leads generally address the 5W1H elements, in opinion articles they are often designed to set the tone or provoke thought through argumentative structures rather than provide factual summarization~\cite{rich1993writing}. Table~\ref{tab:app:comparision_opinion_non-opinion} shows that the model performance varied across the article genres: the F1 score of \textsc{\mymethod} was 0.122 higher on opinion articles than on analysis articles. This trend is consistent with the nature of opinion writing, where stance is often more explicitly expressed. Additionally, we observed genre-specific differences in the effects of ablating segment labels. Removing lead labels caused a more substantial performance drop for analysis articles, and ablating other segment labels also yielded differing effects depending on the article type. We leave further investigation into the sources of these differences to future work.

\begin{table}[t]
\centering
\begin{subtable}[t]{0.8\linewidth}
    \centering
    \resizebox{.8\linewidth}{!}{
    \begin{tabular}{lcc}
    \hline
    \makecell[c]{Model} & ACC & F1 \\\hline
    \makecell[l]{\textsc{\mymethod}} & 0.78 & 0.785\\
    w/o Headline& 0.75 & 0.762\\
    w/o Lead & 0.73 & 0.752\\
    w/o Conclusion & 0.78 & 0.788\\
    w/o Quotations & 0.77& 0.764\\\hline
    \end{tabular}
    }
    \caption{Opinion ($N$=100)}
\end{subtable}

\vspace{2mm}

\begin{subtable}[t]{0.8\linewidth}
    \centering
    \resizebox{.8\linewidth}{!}{
\begin{tabular}{lccc}
    \hline
    \makecell[c]{Model} & ACC & F1 \\\hline
    \makecell[l]{\textsc{\mymethod}} & 0.667 & 0.663\\
    w/o Headline& 0.663 & 0.661\\
    w/o Lead & 0.602 & 0.617\\
    w/o Conclusion & 0.65 & 0.644\\
    w/o Quotations & 0.665 & 0.669\\\hline
    \end{tabular}
    }
    \caption{Analysis ($N$=901)}
\end{subtable}
\caption{Varying performance across different article types when a segment label is ablated for \textsc{\mymethod}.}
\label{tab:app:comparision_opinion_non-opinion}
\end{table}

\paragraph{Trade-offs between accuracy and inference cost}

The proposed \textsc{\mymethod} employed a segment-level LM agent, which led to the improved performance in evaluation experiments. However, this design choice introduced additional computational costs for training the segment-level LM and inferring stance labels. Table~\ref{tab:app:comparison_tradeoff_perf_inf-cost} presents a detailed comparison between the fine-tuned RoBERTa model, LLM zero-shot inference, and the proposed \textsc{\mymethod}. Specifically, our method used six-shot sample augmentation and chain-of-thought prompting, with Gemini-2.0-flash as the LLM backbone. The fine-tuned RoBERTa model achieved the fastest inference time (0.007s per sample) but yielded a lower F1 score of 0.577, which was lower than other models. The LLM zero-shot inference improved the F1 score to 0.628, with an average inference time of 0.599s per sample and an API cost of \$0.0001. The proposed method further increased the F1 score to 0.672, reflecting a notable improvement of 0.095 over the RoBERTa baseline. However, this came at the cost of higher inference latency (1.241s per sample). These results highlight the trade-offs between model performance and inference efficiency: \textsc{\mymethod} offers the highest accuracy among the evaluated methods at the expense of greater computational cost. Nevertheless, the increased inference time is tolerable in practice, as article-level inference can be parallelized.

\begin{table}[t]
\centering
\resizebox{\linewidth}{!}{
\begin{tabular}{ccccc}
\hline
Method & F1 & Training time & Inference time & API cost\\\hline
RoBERTa & 0.577 & 4m 13s & 0.007s & –\\
\makecell[c]{LLM zero-shot} & 0.628 & – & 0.599s & \$0.0001\\
\makecell[c]{\textsc{\mymethod}} & 0.672 & 13m 9s & 1.242s & \$0.0008\\
\hline
\end{tabular}
}
\caption{Accuracy-computation trade-offs. Inference time and API cost are averaged per sample.}
\label{tab:app:comparison_tradeoff_perf_inf-cost}
\end{table}

\begin{table*}[t]
\centering
\resizebox{\linewidth}{!}{
\begin{tabular}{c|cc|cc|cc|cc|cc}
\hline
\multirow{2}{*}{Method} & \multirow{2}{*}{CoT} & \multirow{2}{*}{Few-shot} & \multicolumn{2}{c|}{GPT-4o-mini} & \multicolumn{2}{c|}{Gemini-2.0-flash} & \multicolumn{2}{c|}{Claude-3-haiku} & \multicolumn{2}{c}{EXAONE-2.4b (fine-tuned)}\\\cline{4-11}
& & & Accuracy & F1 & Accuracy & F1 & Accuracy & F1 & Accuracy & F1 \\ \hline
\multicolumn{3}{c|}{\textsc{\mymethod} (RoBERTa)}&  
\textbf{0.607$\pm$0.006} & \textbf{0.602$\pm$0.007} & \textbf{0.678$\pm$0.002}& \textbf{0.672$\pm$0.002}& \textbf{0.639$\pm$0.002} & \textbf{0.638$\pm$0.004}& \textbf{0.601$\pm$0.006} & \textbf{0.599$\pm$0.006} \\\hline
\multirow{4}{*}{\makecell{\textsc{\mymethod}\\(Gemini-2.0-flash)}}&  &  & 0.545$\pm$0.002& 0.549$\pm$0.002 & 0.572$\pm$0.001& 0.549$\pm$0.001 & 0.55$\pm$0.004& 0.526$\pm$0.006 & 0.536$\pm$0.002& 0.529$\pm$0.001 \\ 
& √ &  & 0.56$\pm$0.001& 0.538$\pm$0.002 & \textbf{0.598$\pm$0.001}& \textbf{0.583$\pm$0.002} & 0.56$\pm$0.004 & 0.531$\pm$0.005 & \textbf{0.541$\pm$0.002} & \textbf{0.536$\pm$0.002}\\ 
&  & √ & 0.558$\pm$0.003& 0.539$\pm$0.004 & 0.597$\pm$0.004 & 0.581$\pm$0.005 & \textbf{0.596$\pm$0.004} & \textbf{0.584$\pm$0.004} & 0.325$\pm$0.006& 0.31$\pm$0.007\\ 
& √ & √ & \textbf{0.572$\pm$0.002} & \textbf{0.554$\pm$0.002} & 0.592$\pm$0.003& 0.576$\pm$0.003& 0.589$\pm$0.002& 0.579$\pm$0.003& 0.329$\pm$0.002& 0.304$\pm$0.001 \\ 
\hline
\multirow{4}{*}{\makecell{\textsc{\mymethod}\\(Gemini-2.0-flash\\w/ 6-shot)}}&  &  & 0.548$\pm$0.016& 0.553$\pm$0.016& 0.609$\pm$0.001& 0.587$\pm$0.001& 0.575$\pm$0.001& 0.559$\pm$0.001& \textbf{0.566$\pm$0.001} & \textbf{0.575$\pm$0.001}\\ 
& √ &  & 0.583$\pm$0.008& 0.571$\pm$0.009& 0.625$\pm$0.003& 0.613$\pm$0.004& 0.573$\pm$0.004& 0.547$\pm$0.003& 0.558$\pm$0.002& 0.556$\pm$0.002\\
&  & √ & 0.608$\pm$0.004& 0.598$\pm$0.004& 0.617$\pm$0.003& 0.605$\pm$0.003& \textbf{0.599$\pm$0.004} & \textbf{0.59$\pm$0.004} & 0.343$\pm$0.005& 0.322$\pm$0.004\\
& √ & √ & \textbf{0.611$\pm$0.003} & \textbf{0.602$\pm$0.004}& \textbf{0.63$\pm$0.003} & \textbf{0.617$\pm$0.003} & 0.592$\pm$0.006& 0.587$\pm$0.007& 0.331$\pm$0.004& 0.3$\pm$0.003\\
\hline
\end{tabular}}
\caption{Stance detection performance of \textsc{\mymethod} with LLM agents, measured on the test split of \textsc{\mydata}. The best-performing model for each LLM backbone is marked in bold.}
\label{tab:app:overall_stance_perf_lmm}
\end{table*}

\paragraph{LLM as the segment-level agent}
Table~\ref{tab:app:overall_stance_perf_lmm} presents the performance of two additional variants of \textsc{\mymethod} for predicting the overall stance of news articles. These variants used Gemini-2.0-flash as the segment-level stance detection agent---one with an instruction-only prompt and the other with six-shot samples. Gemini-2.0-flash was selected based on its strong performance on segment-level stance detection (Table~\ref{tab:app:segment_stance_perf}). For comparison, we also report the best performance of \textsc{\mymethod}(RoBERTa) by varying the LLM backbones for article-level stance detection.

The results showed that \textsc{\mymethod} with a fine-tuned RoBERTa agent generally outperformed the LLM-based variants, with the sole exception being GPT-4o-mini, where performance was comparable. Since the best performance across different LLM backbones was achieved by the RoBERTa-based variant, we report its results in Table~\ref{tab:overall_stance_perf}. Nevertheless, considering that LLM-based segment-level agents require few or no labeled examples, the competitive performance of these variants highlights their potential for future investigation.

\paragraph{Segment-level stance detection}

We evaluated the performance of language model agents in predicting stance labels for individual news segments. Specifically, we compared three approaches: fine-tuning a MLM, zero-shot inference with an LLM, and six-shot in-context learning with an LLM. For MLM fine-tuning and few-shot selection, we used the segment-level stance labels and corresponding news text from the training split of \textsc{\mydata}. Table~\ref{tab:app:segment_stance_perf:all} reports the accuracy and macro F1 scores for eight models. 

We found that LLMs generally outperformed fine-tuned RoBERTa models. Given that LLMs require few or no labeled examples, these results highlight their effectiveness for stance detection in short texts. However, despite their strong performance, \textsc{\mymethod} with an LLM agent underperformed in article-level stance detection compared to the RoBERTa-based variant, as shown in Table~\ref{tab:app:overall_stance_perf_lmm}. 

To better understand this discrepancy, we analyzed class-wise performance across different segment types, as shown in Tables~\ref{tab:app:segment_stance_perf:headline} to \ref{tab:app:segment_stance_perf:quotation}. We observed that the RoBERTa model generally performed better in classifying \textit{neutral}-labeled segments, as reflected in higher F1 scores for the \textit{neutral} class. We hypothesize that accurately identifying neutral segments is a key factor contributing to the effectiveness of a segment-level stance agent.

\begin{figure}[t]
    \centerline{\includegraphics[width=.8\columnwidth]{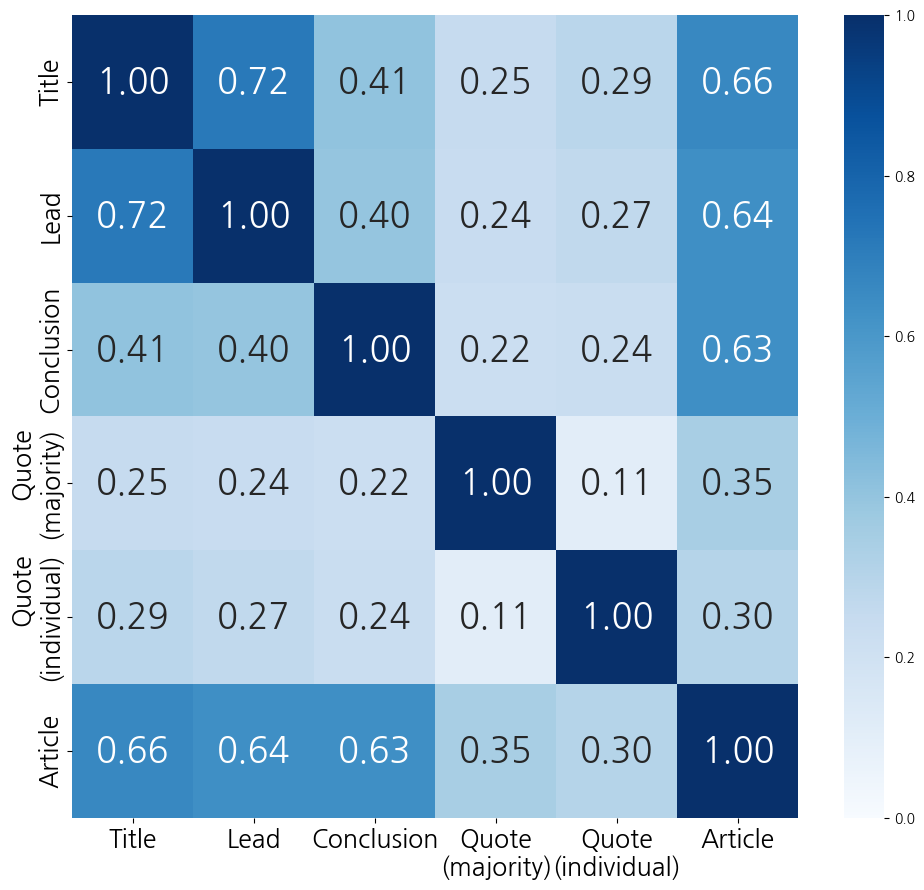}}
    \caption{Inter-segment stance label associations.} 
    \label{fig:label_association} 
\end{figure}

\paragraph{Qualitative error analysis}

From a qualitative analysis of incorrect predictions made by \textsc{\mymethod} using RoBERTa---our best-performing model---we identified two primary error patterns. 

The first arose from model's failure to interpret positive descriptions as indicative of a supportive stance, often leading to the misclassification of supportive articles as neutral. This was the most prominent error type observed in the quantitative error analysis (Figure~\ref{fig:conf_matrix}). For example, one news article expressed a favorable view on the issue of ``Han Dong-hoon: Cutting National Assembly Seats to 250 is on the Table'' by outlining the benefits of the policy proposed by Han. However, the segment-level agent failed to capture this supportive framing, which may subsequently cause the LLM to predict a neutral stance.

The second error pattern emerged during the orchestration of segment-level stance labels. Even when segment-level predictions were accurate, as simulated in the oracle setting, the LLM sometimes failed to infer the correct overall stance. This issue was especially pronounced in articles that contained multiple quotations expressing divergent or conflicting viewpoints.

These two patterns point to potential future directions for improving the article-level stance detection: enhancing the segment-level detection model and selectively incorporating the most salient quotations, rather than considering all of them for label augmentation.

\subsection{Dataset Details}
\label{sec:app:dataset_details}

\begin{table}[h]
    \resizebox{\linewidth}{!}{%
    \begin{tabular}{lccc}
        \hline
        Segment & Supportive (\%) &
        Neutral (\%) &
        Oppositional (\%) \\
        \hline
        Headline & 21.3 & 49.6 & 29.1 \\
        Lead & 20.6 & 52.7 & 26.8 \\
        Conclusion & 27.1 & 41.1 & 31.9 \\
        Quotations & 26.1 & 40.5 & 33.4 \\
        Article & 31.9 & 33.7 & 34.4 \\
        \hline
    \end{tabular}
    }
    \caption{Stance label distribution.}
    \label{tab:stance_dist}
\end{table}

\paragraph{Label distribution} 
Table~\ref{tab:stance_dist} summarizes the distribution of stance labels at both article and segment levels. While the article-level stance labels were relatively balanced across classes, neutral stances appeared more frequently at the segment level. Figure~\ref{fig:label_association} visualizes the relationship between segment-level and article-level stance labels using Cramer's V. Article-level stance showed strong associations with the stances expressed in the headline, lead, and conclusion, each yielding Cramer's V values of approximately 0.7. In contrast, stance labels derived from quotations exhibited weaker associations, with Cramer's V values around 0.3. We also observed a strong correlation between the headline and lead stances, suggesting a shared rhetorical framing established early in the article.

\paragraph{List of target issues and media outlets}
Table~\ref{tab:app:main_dataset_issues} presents a comprehensive list of the target issues covered in \textsc{\mydata}. The dataset includes articles from 31 media outlets, among which the following are the top 10 news agencies: Kyunghyang Shinmun (경향신문), Segye Ilbo (세계일보), Korea JoongAng Daily (중앙일보), Kookmin Ilbo (국민일보), Seoul Shinmun (서울신문), Chosun Daily (조선일보), Seoul Economic Daily (서울경제), Korea Economic Daily (한국경제), MoneyToday (머니투데이), and Hankook Ilbo (한국일보). Table~\ref{tab:app:case_study_dataset_issues} provides the list of target issues that were used in the newly collected dataset for two case studies.

\begin{figure*}[ht]
    \centering{\includegraphics[width=0.77\linewidth]{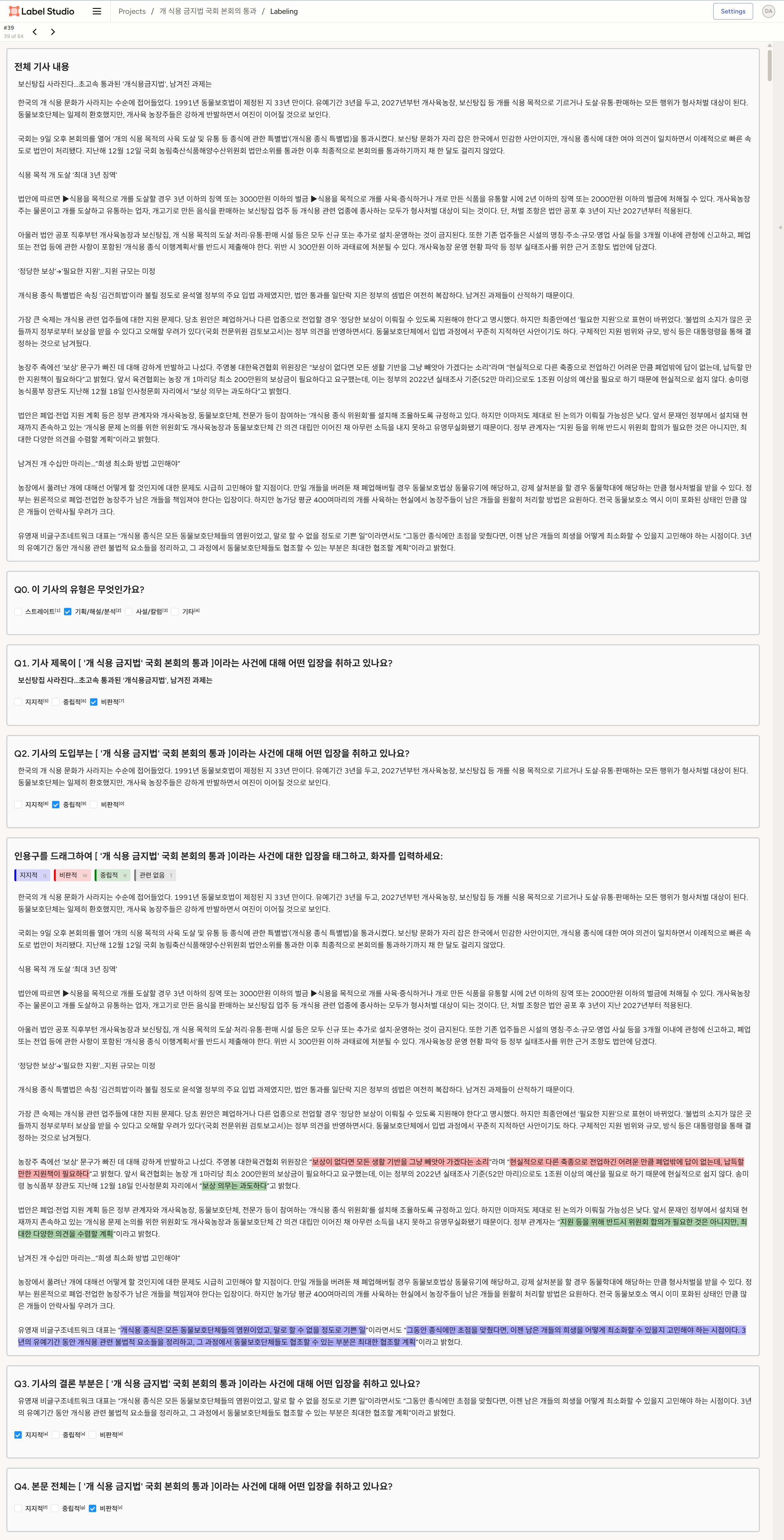}}
    \caption{Labeling interface used during annotation. Both the original and translated guidelines are provided to support future research.}
    \label{fig: labeling_interface} 
\end{figure*}

\paragraph{Dataset access} Instructions for accessing the dataset are available at \url{https://github.com/ssu-humane/K-News-Stance}. Along with the dataset, we provide the annotation guidelines in Korean (Figure~\ref{fig: labeling_interface}) as well as an English-translated version.

\begin{table*}[ht]
\centering
\begin{subtable}[t]{0.9\linewidth}
\centering
\resizebox{\textwidth}{!}{
\begin{tabular}{c|c|cc|ccc}
    \hline
    Type & Model & Accuracy & F1 & \makecell{F1\\(\emph{supportive})} & \makecell{F1\\(\emph{neutral})} & \makecell{F1\\(\emph{oppositional})} \\\hline
  \multirow{2}{*}{\makecell{Fine-tuned\\MLM}}  
    & RoBERTa-base & 0.565$\pm$0.004 & 0.559$\pm$0.006 & 0.442$\pm$0.017 & 0.649$\pm$0.005 & 0.587$\pm$0.011\\
    & RoBERTa-large & \textbf{0.592$\pm$0.006} & \textbf{0.583$\pm$0.01} & \textbf{0.453$\pm$0.031}& \textbf{0.662$\pm$0.005} & \textbf{0.634$\pm$0.011} \\ \hline
  \multirow{3}{*}{\makecell{LLM\\(zero-shot)}} & GPT-4o-mini & 0.592$\pm$0.002& 0.578$\pm$0.003& 0.476$\pm$0.001& 0.570$\pm$0.002& 0.687$\pm$0.004
\\
    & Gemini-2.0-flash & 
\textbf{0.647$\pm$0.004}& \textbf{0.638$\pm$0.003}& \textbf{0.575$\pm$0.003}& \textbf{0.594$\pm$0.002}& \textbf{0.745$\pm$0.001}
\\
    & Claude-3-haiku & 0.587$\pm$0.001& 0.58$\pm$0.004&  0.49$\pm$0.003&  0.542$\pm$0.002& 0.71$\pm$0.003\\\hline
    \multirow{3}{*}{\makecell{LLM\\(6-shot)}}& GPT-4o-mini & 
0.598$\pm$0.004& 0.583$\pm$0.004& 0.469$\pm$0.004& 0.574$\pm$0.003& 0.707$\pm$0.004\\
    & Gemini-2.0-flash & \textbf{0.671$\pm$0.004}& \textbf{0.664$\pm$0.004}& 0.6$\pm$0.004& \textbf{0.623$\pm$0.003}& \textbf{0.771$\pm$0.004}
\\
    & Claude-3-haiku & 0.643$\pm$0.002& 0.636$\pm$0.002& \textbf{0.613$\pm$0.003}& 0.534$\pm$0.003& 0.761$\pm$0.001\\\hline
\end{tabular}
}
\caption{Performance on all segments}
\label{tab:app:segment_stance_perf:all}
    \vspace{2mm}
\end{subtable}

\begin{subtable}[t]{0.9\linewidth}
\centering
\resizebox{\textwidth}{!}{
\begin{tabular}{c|c|cc|ccc}
    \hline
    Type & Model & Accuracy & F1 & \makecell{F1\\(\emph{supportive})} & \makecell{F1\\(\emph{neutral})} & \makecell{F1\\(\emph{oppositional})} \\\hline
  \multirow{2}{*}{\makecell{Fine-tuned\\MLM}}  
    & RoBERTa-base & 0.636$\pm$0.005 & 0.576$\pm$0.008 & 0.428$\pm$0.021 & 0.719$\pm$0.005 & 0.581$\pm$0.011\\
    & RoBERTa-large & \textbf{0.664$\pm$0.006} & \textbf{0.61$\pm$0.008} & \textbf{0.45$\pm$0.021} & \textbf{0.731$\pm$0.007} & \textbf{0.648$\pm$0.02}\\ \hline
  \multirow{3}{*}{\makecell{LLM\\(zero-shot)}} & GPT-4o-mini & 0.65$\pm$0.004& 0.636$\pm$0.003& \textbf{0.577$\pm$0.004}& \textbf{0.683$\pm$0.004}& 0.647$\pm$0.001
\\
    & Gemini-2.0-flash & \textbf{0.658$\pm$0.004}& \textbf{0.642$\pm$0.005}& 0.563$\pm$0.002& 0.68$\pm$0.002& \textbf{0.682$\pm$0.003}
\\
    & Claude-3-haiku & 0.623$\pm$0.004& 0.597$\pm$0.003&  0.469$\pm$0.003&  0.668$\pm$0.004& 0.655$\pm$0.004\\\hline

    \multirow{3}{*}{\makecell{LLM\\(6-shot)}}& GPT-4o-mini & 0.656$\pm$0.002& 0.636$\pm$0.004& 0.554$\pm$0.003& 0.695$\pm$0.003& 0.659$\pm$0.002\\
    & Gemini-2.0-flash & \textbf{0.702$\pm$0.002}& \textbf{0.687$\pm$0.004}& \textbf{0.603$\pm$0.003}& \textbf{0.724$\pm$0.003}& \textbf{0.733$\pm$0.002}\\
    & Claude-3-haiku & 0.64$\pm$0.003& 0.635$\pm$0.004& 0.56$\pm$0.001& 0.619$\pm$0.003& 0.727$\pm$0.001\\\hline
\end{tabular}
}
\caption{Performance on \textsf{Headline}}
\label{tab:app:segment_stance_perf:headline}
    \vspace{2mm}
\end{subtable}

\begin{subtable}[t]{0.9\linewidth}
\centering
\resizebox{\textwidth}{!}{
\begin{tabular}{c|c|cc|ccc}
    \hline
    Type & Model & Accuracy & F1 & \makecell{F1\\(\emph{supportive})} & \makecell{F1\\(\emph{neutral})} & \makecell{F1\\(\emph{oppositional})} \\\hline
  \multirow{2}{*}{\makecell{Fine-tuned\\MLM}}  
    & RoBERTa-base & 0.675$\pm$0.007 & \textbf{0.61$\pm$0.008} & 0.465$\pm$0.014 & 0.752$\pm$0.008 & \textbf{0.612$\pm$0.019}\\
    & RoBERTa-large & \textbf{0.681$\pm$0.006} & 0.608$\pm$0.012 & \textbf{0.475$\pm$0.018} & \textbf{0.758$\pm$0.004} & 0.592$\pm$0.029\\ \hline
  \multirow{3}{*}{\makecell{LLM\\(zero-shot)}} & GPT-4o-mini & 0.668$\pm$0.005& 0.62$\pm$0.004& 0.518$\pm$0.004& \textbf{0.74$\pm$0.002}& 0.601$\pm$0.003
\\
    & Gemini-2.0-flash & \textbf{0.677$\pm$0.002}& \textbf{0.648$\pm$0.002}& \textbf{0.554$\pm$0.002}& 0.727$\pm$0.002& \textbf{0.662$\pm$0.003}
\\
    & Claude-3-haiku & 0.581$\pm$0.003& 0.548$\pm$0.003&  0.387$\pm$0.002&  0.641$\pm$0.001& 0.616$\pm$0.003\\\hline
    
    \multirow{3}{*}{\makecell{LLM\\(6-shot)}}& GPT-4o-mini & 0.676$\pm$0.004& 0.634$\pm$0.002& 0.499$\pm$0.001& 0.741$\pm$0.001& 0.661$\pm$0.005\\
    & Gemini-2.0-flash & \textbf{0.69$\pm$0.003} & \textbf{0.664$\pm$0.002}& \textbf{0.547$\pm$0.003} & 0.743$\pm$0.002 & \textbf{0.7$\pm$0.001}\\
    & Claude-3-haiku & 0.637$\pm$0.005& 0.636$\pm$0.004& 0.534$\pm$0.007 & \textbf{0.761$\pm$0.007}& 0.613$\pm$0.003\\\hline
\end{tabular}
}
\caption{Performance on \textsf{Lead}}
\label{tab:app:segment_stance_perf:lead}
    \vspace{2mm}
\end{subtable}

\begin{subtable}[t]{0.9\linewidth}
\centering
\resizebox{\textwidth}{!}{
\begin{tabular}{c|c|cc|ccc}

    \hline
    Type & Model & Accuracy & F1 & \makecell{F1\\(\emph{supportive})} & \makecell{F1\\(\emph{neutral})} & \makecell{F1\\(\emph{oppositional})} \\\hline
  \multirow{2}{*}{\makecell{Fine-tuned\\MLM}}  
    & RoBERTa-base & 0.576$\pm$0.005 & 0.546$\pm$0.005 & 0.422$\pm$0.019 & 0.63$\pm$0.01 & 0.586$\pm$0.012\\
    & RoBERTa-large & \textbf{0.594$\pm$0.006} & \textbf{0.559$\pm$0.01} & \textbf{0.431$\pm$0.025} & \textbf{0.65$\pm$0.005} & \textbf{0.597$\pm$0.018}\\ \hline
  \multirow{3}{*}{\makecell{LLM\\(zero-shot)}} & GPT-4o-mini & 0.634$\pm$0.002& 0.603$\pm$0.005& 0.447$\pm$0.004& 0.675$\pm$0.004& 0.686$\pm$0.003
\\
    & Gemini-2.0-flash & \textbf{0.681$\pm$0.003}& \textbf{0.661$\pm$0.004}& \textbf{0.545$\pm$0.004}& \textbf{0.703$\pm$0.002}& \textbf{0.735$\pm$0.004}
\\
    & Claude-3-haiku & 0.592$\pm$0.002& 0.576$\pm$0.003&  0.457$\pm$0.004&  0.598$\pm$0.002& 0.671$\pm$0.001\\\hline
    \multirow{3}{*}{\makecell{LLM\\(6-shot)}}& GPT-4o-mini & 0.635$\pm$0.005& 0.602$\pm$0.004& 0.431$\pm$0.002& \textbf{0.667$\pm$0.005}& 0.708$\pm$0.004\\
    & Gemini-2.0-flash & \textbf{0.66$\pm$0.005}& \textbf{0.65$\pm$0.004}& \textbf{0.56$\pm$0.002}& 0.65$\pm$0.005& \textbf{0.74$\pm$0.004}\\
    & Claude-3-haiku & 0.6$\pm$0.002& 0.593$\pm$0.005& \textbf{0.56$\pm$0.003} & 0.625$\pm$0.002& 0.593$\pm$0.003\\\hline
\end{tabular}
}
\caption{Performance on \textsf{Conclusion}}
\label{tab:app:segment_stance_perf:conclusion}
    \vspace{2mm}
\end{subtable}

\begin{subtable}[t]{0.9\linewidth}
\centering
\resizebox{\textwidth}{!}{

\begin{tabular}{c|c|cc|ccc}
    \hline
    Type & Model & Accuracy & F1 & \makecell{F1\\(\emph{supportive})} & \makecell{F1\\(\emph{neutral})} & \makecell{F1\\(\emph{oppositional})} \\\hline
  \multirow{2}{*}{\makecell{Fine-tuned\\MLM}}  
    & RoBERTa-base & 0.541$\pm$0.006 & 0.536$\pm$0.007 & 0.443$\pm$0.02 & 0.582$\pm$0.006 & 0.582$\pm$0.01\\
    & RoBERTa-large & \textbf{0.571$\pm$0.01} & \textbf{0.563$\pm$0.013} & \textbf{0.452$\pm$0.037} & \textbf{0.591$\pm$0.006} & \textbf{0.647$\pm$0.006}\\ \hline
  \multirow{3}{*}{\makecell{LLM\\(zero-shot)}} & GPT-4o-mini & 0.57$\pm$0.004& 0.552$\pm$0.002& 0.465$\pm$0.005& 0.494$\pm$0.005& 0.697$\pm$0.004
\\
    & Gemini-2.0-flash & \textbf{0.636$\pm$0.002}& \textbf{0.622$\pm$0.001}& \textbf{0.58$\pm$0.004}& \textbf{0.524$\pm$0.004}& \textbf{0.761$\pm$0.004}
\\
    & Claude-3-haiku & 0.582$\pm$0.004& 0.573$\pm$0.003&  0.504$\pm$0.004&  0.486$\pm$0.002& 0.729$\pm$0.001\\\hline
    \multirow{3}{*}{\makecell{LLM\\(6-shot)}}& GPT-4o-mini & 0.575$\pm$0.003& 0.557$\pm$0.002& 0.461$\pm$0.002& 0.494$\pm$0.003& 0.715$\pm$0.002\\
    & Gemini-2.0-flash & \textbf{0.669$\pm$0.003}& \textbf{0.658$\pm$0.002}& 0.608$\pm$0.002& \textbf{0.581$\pm$0.003}& 0.786$\pm$0.002
\\
    & Claude-3-haiku & 0.649$\pm$0.003& 0.64$\pm$0.003& \textbf{0.614$\pm$0.005}& 0.5$\pm$0.003& \textbf{0.805$\pm$0.003}\\\hline
\end{tabular}

}
\caption{Performance on \textsf{Quotations}}
\label{tab:app:segment_stance_perf:quotation}
    \vspace{2mm}
\end{subtable}

\caption{Segment-level stance detection performance, measured on the test split of \textsc{\mydata}.}
\label{tab:app:segment_stance_perf}
\end{table*}

\begin{table*}[ht]
    \centering
    \scriptsize
    \begin{tabularx}{\linewidth}{c|X|X}
    \hline
        Date & Issue (in Korean) &  Issue (in English) \\\hline
       2022-06-15 & 화물연대 파업 8일만에 철회, 안전운임제는 지속 추진  &  Truckers' Strike Ends After 8 Days, Safety Freight Rates to Continue \\
       2022-06-16 & 내년 최저임금 업종 구분 없이 동일 적용 & Next Year's Minimum Wage to be Applied Uniformly Regardless of Industry \\
       2022-06-22& 윤석열 정부 탈원전 폐기 공식화 & Yoon Administration Officially Ends Nuclear Phase-Out Policy \\
       2022-07-17 & 이재명 당대표 선거 출마 선언 & Lee Jae-myung Announces Candidacy for Party Leadership Election \\
       2022-07-31 & 초등 입학연령 하향 추진 & Elementary school entry age to be lowered to five \\
       2022-08-10 & 박민영 대통령실 청년대변인 발탁 & Park Min-young Appointed as Presidential Spokesperson for Youth Affairs \\
       2022-10-06 & 정부 여성가족부 폐지 정부조직개편안 확정 & Government confirms organizational restructuring plan to abolish Ministry of Gender Equality and Family \\ 
       2022-10-11 & 정부 학업성취도 자율평가 확대 추진 & Government Revives and Expands Autonomous Academic Evaluation Testing \\
       2022-10-26 & 법무부, 촉법소년 기준 만 13세로 1년 하향 & Ministry of Justice Lowers Age Threshold for Juvenile Offenders to 13 \\
       2022-11-07 & 문재인 대통령, 키우던 풍산개 2마리 정부 반환 & Moon Jae-in Hands Over Two North Korean Pungsan Dogs to the State \\
       2022-12-19 & 국민의힘, 당원투표 100\% 결선투표제 도입 & People Power Party approves 100\% party-member vote with a runoff system for leadership election \\
       2022-12-21 & 정부, 다주택자 부동산 규제 대폭 완화 & Government Eases Real Estate Regulations for Multiple Homeowners \\
       2022-12-21 & 윤석열 대통령 ``노조부패, 척결해야할 3대 부패'' & President Yoon Calls ``Union Corruption is One of Three Major Evils to Be Eradicated'' \\
       2023-01-12 & 정부 `강제징용 배상 일본 기업 대신 지급' 공식화 & Government to compensate forced-labor victims on behalf of Japanese firms \\
       2023-02-15 & 노란봉투법 국회 상임위 통과 & Labor-Friendly `Yellow Envelope' labor bill passes National Assembly standing committee \\
       2023-02-21 & 법원 동성 부부 배우자도 건강보험 피부양자 인정 & Court Recognizes Same-Sex Partners as Legal Dependents for Health Insurance \\
       2023-03-06 & 주 52시간 근로시간 개편 최대 69시간 가능 & Government Proposes Overhaul of 52-Hour Workweek, Allowing Up to 69 Hours \\
       2023-03-23 & 헌재 `검수완박' 절차는 위헌, 법안은 유효 & Constitutional Court Upholds Prosecutorial Reform Laws, Acknowledges Procedural Violations \\
       2023-03-23 & `양곡관리법 개정안' 민주당 주도 본회의 통과 & Democratic Party-Led Revision of the Grain Management Act Put Directly to Plenary Session for Vote \\
       2023-04-05 & 학폭 가해기록 보존기간 연장·입시반영 검토 & School Bullying Records May Be Kept Longer, Reflected in College Admissions \\
       2023-05-22 & 등록 재산에 가상자산 포함법안 통과 & Bill Passed to Include Cryptocurrency in Public Asset Declarations \\
       2023-06-18 & 당정, 중대 범죄자 신상공개 특별법 추진 & Ruling Party and Government Push Special Bill to Disclose Identities of Serious Criminals \\
       2023-06-28 & `출생통보제' 법사위 소위 통과 & `Birth Notification System Bill' Advances in Parliament \\
       2023-07-04 & IAEA 보고서 일본 오염수 방류 문제 없다 & IAEA Report Says Japan's Fukushima Water Release Poses No Safety Concerns \\
       2023-07-12 & 정부·여당, 실업급여 하한액 낮추거나 폐지까지 검토 & Government and Ruling Party Consider Lowering or Abolishing Minimum Unemployment Benefit \\
       2023-07-20 & 환경부 4대강 16개보 모두 존치 & All 16 Weirs on Four Rivers to Remain Intact, Says Environment Ministry \\
       2023-07-31 & 정부, 외국인 가사도우미 100여명 시범 도입 & Seoul to Test Foreign Domestic Worker Program with Initial 100 Hires\\
       2023-08-22 & 새 대법원장 후보에 이균용 서울고법 부장판사 & Lee Kyun-yong, Nominated as New Chief Justice Candidate \\
       2023-08-23 & 한덕수 총리 ``흉악 범죄 예방 위해 의경 재도입 검토'' & Prime Minister Han Duck-soo Considers Reintroducing Conscripted Police to Prevent Heinous Crimes \\
       2023-09-26 & 헌재 `대북 전단 금지법' 위헌 결정 & Constitutional Court Invalidates Law Banning Leaflets to North Korea \\
       2023-10-12 & 검찰 이재명 `백현동 배임혐의' 불구속 기소 & Prosecution Indicts Lee Jae-myung Without Detention over Baekhyeon-dong Scandal \\
       2023-11-02 & 국민의힘 `김포 서울 편입' 특위 발족 & People Power Party Launches Special Committee for Gimpo-Seoul Integration \\
       2023-11-05 & 내년 6월까지 공매도 전면 금지 & Short Selling Fully Banned Until June 2025 \\
       2023-12-07 & 민주당, 전당대회서 권리당원 표 비중 확대 & DP Boosts Role of Rank-and-File Members in Party Convention Votes \\
       2023-12-21 & 금리 4\% 넘는 자영업자 최대 300만 원 환급 & Business Owners Paying Over 4\% Interest Eligible for Government Refunds \\
       2023-12-27 & 이준석 국민의힘 탈당 후 신당 창당 돌입 & Lee Jun-seok Leaves People Power Party and Begins Forming New Party \\
       2024-01-16 & 한동훈 ``국회의원 250명으로 줄이겠다'' & Han Dong-hoon: Cutting National Assembly Seats to 250 is on the Table\\
       2024-01-29 & 이준석 ``경찰·소방관 되려는 여성 군 복무해야'' & Lee Jun-seok: Women Should Serve in Military to Join Police or Firefighting Forces \\
       2024-01-09 & `개 식용 금지법' 국회 본회의 통과 & National Assembly Passes Bill Banning Dog Meat Consumption \\
       2024-02-06 & 내년 의대 정원 2천 명 증원 & Medical School Admissions to Increase by 2,000 Next Year \\
       2024-02-13 & 조국 전 법무부 장관 신당 창당 선언 & Former Justice Minister Cho Kuk Declares Launch of New Party \\
       2024-03-27 & 한동훈 국회 세종시로 완전 이전 공약 발표 & Justice Minister Han Dong-hoon Proposes Relocating the National Assembly to Sejong City \\
       2024-04-24 & 한강에서 먹고 자고 일한다 `리버시티 서울' 발표 & Seoul Announces `River City' Project to Create Living, Working, and Leisure Spaces on the Han River \\
       2024-04-10 & 경기 화성을 개혁신당 이준석 대표 당선 & Reform Party Leader Lee Jun-seok Wins Parliamentary Seat in Hwaseong, Gyeonggi \\
       2024-05-08 & `외국 면허 의사' 국내 진료 허용 & Foreign-Licensed Doctors Allowed to Practice in Korea \\
       2024-05-30 & 법원 하이브 `민희진 해임' 제동 & Court Rejects Hive's Attempt to Dismiss Min Hee-jin \\
       2024-06-03 & 윤석열 대통령 ``동해 140억배럴 석유·가스 매장 추정'' & President Yoon: East Sea May Hold 14 Billion Barrels of Oil and Gas \\\hline
    \end{tabularx}
    \caption{A comprehensive list of target issues in \textsc{\mydata}.}
    \label{tab:app:main_dataset_issues}
\end{table*}

\begin{table*}[ht]
    \centering
    \scriptsize
    \begin{tabularx}{\linewidth}{c|c|X|X}
    \hline
        Study & Date & Issue (in Korean) &  Issue (in English) \\\hline
        \multirow{8}{*}{Recommendation} & 2024-07-21 & 군, 북한 오물풍선 잇단 살포에 대북 방송 전면 시행 & Military to Resume Full-Scale Propaganda Broadcasts to North Following Repeated Trash Balloon Incidents \\
       &2024-08-12 & 윤석열 대통령, 검찰총장 후보자 심우정 법무장관 지명 & President Yoon Nominates Vice Justice Minister Shim Woo-jung for Prosecutor General \\
       &2024-12-03 & 사상초유 감사원장 탄핵안 내일 표결 & Unprecedented Impeachment Motion Against Board of Audit and Inspection Chairman Faces Vote Tomorrow \\
       &2024-12-27 & 민주, 한 권한대행 탄핵안 발의...27일 표결 & Minjoo Party Files Impeachment Motion Against Acting President Han; Vote Set for 27th \\ \hline
       \multirow{3}{*}{Media bias}&2024-12-26 & 한덕수 권한대행 탄핵안 본회의 통과 & Acting President Han Duck-soo Impeachment Passes Parliament \\
       &2025-04-30 & 대법원, 이재명 대선후보 공직선거법 사건 파기환송 & Supreme Court Remands Lee Jae-myung's Election Law Violation Case \\\hline
       \end{tabularx}
    \caption{List of target issues covered in the two case studies.}
    \label{tab:app:case_study_dataset_issues}
\end{table*}

\begin{table*}[t]
    \centering
    \small
        \begin{subtable}[b]{\linewidth}
    \begin{tabular}{p{15cm}}
         \makecell{\normalsize\textbf{Target Issue}: `개 식용 금지법' 국회 본회의 통과}\\\hline
         \textbf{Headline} (\emph{Supportive}): `개 식용' 역사 끝날까···동물권단체 `개 식용 금지 특별법' 통과에 \colorbox{cyan!30}{``실질적 종식 기대''}\\
         \textbf{Body Text}\\
         -\textsf{Lead} (\emph{Supportive}): 동물권단체들 ‘개 식용 금지법’ 국회 통과에 일제히 환영. \colorbox{cyan!30}{``실질적으로 개 식용 금지 효과,} \colorbox{cyan!30}{ 가까운 시일 내 종식 기대.''} 육견협회는 강하게 반발 \colorbox{red!30}{``1000만 국민의 먹을권리 빼앗아.''} `개의 식용 목적의 사육·도살 및 유통 등 종식에 관한 특별법 제정안'이 9일 국회 본회의를 통과했다. 그동안 개 식용 금지를 위한 활동을 벌여온 동물권단체들은 \colorbox{cyan!30}{``특별법 통과가 개 식용 없는 나라로 가는 첫걸음이 될 것''}이라며 환영했다.\\
         -\textsf{Conclusion} (\emph{Neutral}): 앞서 육견협회는 지난해 11월 서울 용산 대통령실 인근에서 개식용금지법에 항의하며 개 200만 마리를 서울 일대에 풀겠다고 예고하기도 했다.\\
         \textbf{Overall Stance}: \emph{\textcolor{blue}{Supportive}}\\
         \hline
    \end{tabular}%
    \end{subtable}   
    \begin{subtable}[b]{\linewidth}
    \begin{tabular}{p{15cm}}
         \textbf{Headline} (\emph{Neutral}): 
         ‘개 식용 금지법’ 통과에 \colorbox{cyan!30}{``기념비적 역사''} vs \colorbox{red!30}{``헌법소원 낼 것''}\\
         \textbf{Body Text}\\
         -\textsf{Lead} (\emph{Neutral}): 
         동물단체-육견협회, 엇갈린 반응. \colorbox{cyan!30}{``동물권 승리''} vs \colorbox{red!30}{``먹을 권리 강탈''}. 이른바 '개 식용 금지법'이 9일 국회를 통과하자 동물단체들은  \colorbox{cyan!30}{``기념비적인 역사가 쓰였다''}며 일제히 환영했다. 반면 이를 줄곧 반대해 온 대한육견협회는 \colorbox{red!30}{``직업선택의 자유를 빼앗았다''}며 헌법소원을 내겠다는 뜻을 밝혔다.\\
         -\textsf{Conclusion} (\emph{Neutral}): 이날 국회 본회의에서 의결된 '개의 식용 목적의 사육·도살 및 유통 등 종식에 관한 특별법'은 식용 목적의 개 도살·사육·증식, 개나 개를 원료로 한 식품의 유통·판매를 금지하는 것을 골자로 한다.\\
         \textbf{Overall Stance}: \textcolor{teal}{\emph{Neutral}}\\
         \hline
    \end{tabular}
    \end{subtable}
    \begin{subtable}[b]{\linewidth}
    \begin{tabular}{p{15cm}}
         \parbox[t]{\linewidth}{
         \raggedright
         \textbf{Headline} (\emph{Oppositional}): 개 식용 금지법 통과…\colorbox{red!30}{``20년 보신탕 팔았는데 살길 막막''}\\
         \textbf{Body Text}\\
         -\textsf{Lead} (\emph{Oppositional}): 관련업주 당혹, 2027년부터 처벌. 처벌 수위·적정성 놓고 논란 제기. 사육·유기견 급증 해결 급선무. 개 식용 금지법이 9일 국회 본회의를 통과하면서 개 식용 논쟁이 다시 불거졌다. 정부는 이번 법안을 통해 더이상의 논란을 막겠다는 입장이지만 처벌 수위, 처벌의 적정성을 두고 또 다른 논란이 제기되고 있다.\\
         -\textsf{Conclusion} (\emph{Oppositional}):
         춘천에서 20년 간 영양탕집을 운영한 A씨는 \colorbox{red!30}{``20년 간 장사했는데 금지되면 어떤 업종} \colorbox{red!30}{을 해야 될 지 막막하다.''} \colorbox{red!30}{``더 한 것도 먹으면서 왜 갑자기 보신탕을 금지시키느냐.}\colorbox{red!30}{자꾸 우리를 공격하니까 답답하}\\\colorbox{red!30}{다''}고 했다. 또다른 업주 B씨는 \colorbox{red!30}{``작업장이 싹 없어져 고기 구할 곳도 없어 문을 닫고 쉰 적도 있다''}며 \colorbox{red!30}{``임대료가 한}\\ \colorbox{red!30}{ 달에 160만원 나가서 하루에 80만원은 팔아야 남는데 요즘 같은 경우는 20만원 겨우 팔고 있어 월세도 못 내는 형}\\\colorbox{red!30}{편''}이라고 토로했다. 법안 통과로 인해 처벌 수위와 적정성에 대한 논란이 재점화되었고, 사육 중인 개의 급증과 유기견 문제 해결도 시급하다는 지적이 나왔다.\\}
         \textbf{Overall Stance}: \emph{\textcolor{red}{Oppositional}}\\
         \hline
    \end{tabular}
    \end{subtable}
    \caption{The original example in Korean. The colored highlight indicates the stance label for quotations (blue: supportive, red: oppositional).}
    \label{tab:labeling_example_korean_original}
\end{table*}

\subsection{Used Prompts}

Figures~\ref{fig:proposed_prompt} and \ref{fig:proposed_prompt_korean_original} show the prompts used for article-level stance detection with an LLM: the English-translated version and the original Korean version, respectively. For training the RoBERTa model employed as the segment-level stance prediction agent, we used the following input template: $\texttt{[CLS]} \texttt{ issue } \texttt{[SEP]} \texttt{ segment } \texttt{[SEP]}$.

\begin{figure*}[ht]
\begin{tcolorbox}[colback=white, fontupper=\small]
\textbf{[System Prompt]}\\
Stance detection is the task of determining the expressed or implied opinion, or stance, of a statement toward a certain, specified target. You are given an issue and a news article about that issue. Your task is to classify the article's stance toward the given issue as one of the following: supportive, neutral, or oppositional.\\

The criteria for each label are as follows: \\
- Supportive: The article shows a favorable tone toward the issue, emphasizes quotations in support of the issue, and predominantly uses positive or optimistic language.\\
- Neutral: The article maintains an objective tone, balances quotations from both supportive and critical perspectives, and uses neutral language.\\
- Oppositional: The article shows a skeptical tone toward the issue, \\emphasizes quotations that criticize the issue, and predominantly uses negative or pessimistic language.\\

Additional information is provided on the stance of the headline, lead, conclusion, and quotations regarding the issue.\\
Each segment is marked with XML tags, and the final stance should be determined by taking into account the detailed stance labels of each part.\\

\textbf{[User Prompt]}\\
Issue: \textit{\textcolor{blue}{Government confirms organizational restructuring plan to abolish Ministry of Gender Equality and Family}}\\\\
Headline: \textit{\textcolor{blue}{<Headline stance=``Oppositional''>MOGEF downgraded to a department... Concerns ``Gender equality policies will be buried in the giant MOHW''</Headline>}}\\\\
Article: \textit{\textcolor{blue}{
<Lead stance=``Oppositional''>Under the government restructuring plan announced on the 6th, the Ministry of Gender Equality and Family (MOGEF) faces demotion to a department under the MOHW after 21 years as an independent ministry. The government emphasizes that MOGEF’s functions will be retained and may create synergy with the MOHW’s welfare policy capabilities. Even experts who support the reorganization question whether the giant MOHW can respond quickly to gender equality issues.</Lead>\\\\
MOGEF highlights that integrating its youth policies with MOHW’s child welfare functions can yield synergistic effects. On this day, Minister Kim Hyunsook announced a `Support Plan for In- and Out-of-School Youth' and said, <Quotation stance=``Neutral''>If we became a department with substantial authority under the MOHW, we could have included more in today’s announcement</Quotation> According to the Ministry of the Interior and Safety’s restructuring plan, functions like support for career-interrupted women will be transferred to the Ministry of Employment and Labor, The four core functions—△youth, △family, △women and gender equality, and △rights (e.g., support for victims of sexual/domestic violence)—will be transferred to the Population, Family, and Gender Equality Bureau under the MOHW. Some argue that organizations led by ministers and those led by department heads have significantly different authority within government. Huh Min-sook, a legislative researcher at the National Assembly, said, <Quotation stance=``Oppositional''>MOGEF already lacked budget and authority, making cooperation difficult — demoting it will only weaken it further</Quotation> \\\\
There are concerns that the control tower responsible for formulating gender equality policies and overseeing their implementation across all government agencies will disappear. Park Sun-young, senior researcher at the Korean Women’s Development Institute, stated, <Quotation stance=``Oppositional''>Gender equality policy is about coordination across all ministries — that’s why MOGEF was created,</Quotation> and pointed out that <Quotation stance=``Oppositional''>putting it under the implementation-focused MOHW would undermine its effectiveness.</Quotation> Even experts who criticize MOGEF’s performance say transferring and downsizing its functions to the MOHW would hinder gender equality policy. Jung Jae-hoon, professor at Seoul Women’s University, noted <Quotation stance=``Supportive''>a ``gender ghetto'' phenomenon had emerged in which gender issues were discussed only among women within MOGEF</Quotation> and stated that <Quotation stance=``Supportive''>a presidential committee on gender equality should be established to elevate the issue to the level of the President’s agenda</Quotation>. Hong Sung-geol, professor of public administration at Kookmin University, stated that <Quotation stance=``Supportive''>in the case of family policy, separating it from the MOHW’s welfare agenda had led to fragmented policy momentum</Quotation>and viewed the restructuring positively. However, he also stated, <Quotation stance=``Oppositional''>A gender equality committee that can evaluate all ministries’ policies is the ideal approach</Quotation>\\\\
<Conclusion Stance=``Oppositional''>Within MOGEF, concerns are rising that the policies will be treated as secondary if placed under the MOHW. One MOGEF official said, <Quotation stance=``Oppositional''>Our role is to protect those who cannot raise their voices, based on awareness of diversity and gender</Quotation>and added, <Quotation stance=``Oppositional''>These duties are bound to become insignificant within the vast MOHW</Quotation></Conclusion>}}
\end{tcolorbox}
\centering
\caption{The English-translated prompt used in \textsc{\mymethod}, shown with an illustrative input. Blue italic text highlights the input.}
\label{fig:proposed_prompt}
\end{figure*}

\begin{figure*}[ht]
\begin{tcolorbox}[colback=white, fontupper=\small]
\textbf{[System Prompt]}\\
입장 분류는 특정 대상에 대한 텍스트의 명시적 또는 묵시적인, 의견이나 입장을 결정하는 작업입니다. 이슈와 뉴스 기사가 제공되며, 당신의 임무는 주어진 이슈에 대한 뉴스 기사의 입장을 지지적, 중립적 혹은 비판적 중 하나로 분류하는 것입니다.\\

각 라벨의 판단 기준은 다음과 같습니다: \\
- 지지적: 이슈에 대해 호의적인 논조, 옹호하는 입장의 인용문을 중심으로 배치하며, 긍정적·낙관적 어조가 지배적인 경우\\
- 중립적: 이슈에 대해 객관적인 논조, 옹호하거나 비판하는 입장의 인용문을 균형 있게 배치하며, 중립적 어조를 사용하는 경우\\
- 비판적: 이슈에 대해 회의적인 논조, 비판하는 입장의 인용문을 중심으로 배치하며, 부정적·비관적 어조가 지배적인 경우\\

추가 정보로 이슈에 대한 제목, 도입부, 결론부, 직접인용구의 입장 정보가 각각 제공됩니다. \\
각 위치는 XML 태그로 표시되며, 세부 라벨 정보를 함께 고려하여 최종 입장을 결정하세요. \\

\textbf{[User Prompt]}\\
이슈: \textit{\textcolor{blue}{정부 여성가족부 폐지 정부조직개편안 확정}}\\\\
제목: \textit{\textcolor{blue}{<제목 입장=``비판적''>`본부'로 격하된 여가부...``공룡 복지부에서 성평등 정책 묻힐 것'' 우려</제목>}}\\\\
기사: \textit{\textcolor{blue}{
<도입부 입장=``비판적''> 6일 발표된 정부조직 개편안에 따라 여성가족부가 출범 21년 만에 독립부처에서 보건복지부 산하 본부로 격하될 위기에 처했다. 정부는 여가부 기능은 유지되고, 보건복지부의 복지 정책 기능과 시너지를 낼 수 있다는 점을 강조한다. 그러나 여가부 개편에 찬성하는 전문가들도 `공룡 부처'가 되는 보건복지부가 성평등 문제에 기민하게 대처할 수 있을지 의문을 제기하고 있다.</도입부>\\\\
여가부는 복지부의 아동복지 기능과 여가부의 청소년 정책 등이 통합되면  시너지 효과를 낼 수 있다는 점을 강조하고 있다. 김현숙 여가부 장관은 이날 '학교 안팎 청소년 지원 강화 대책'을 발표하며 <직접인용구 입장=``중립적''>복지부 산하의 상당한 권한을 가진 본부가 된다면 오늘 여가부가 발표한 내용보다 더 많은 내용이 담길  수 있었을 것</직접인용구>이라고 말했다. 
행정안전부의 정부조직 개편안에 따르면 기존 여가부 기능 중 경력단절여성 지원 등 여성고용 기능은 고용노동부로 이관하고, △청소년 △가족 △여성 및 성평등 △권익(성폭력, 가정폭력 등 피해자 지원) 등 4대 기능은 복지부 산하 인구가족양성평등본부로  이관된다.\\\\
그러나 장관이 이끄는 조직과 본부장이 이끄는 조직이 정부 부처 내에서 갖는 위상이 다르다는 주장이 나온다. 허민숙 국회 입법조사관은<직접 인용구 입장=``비판적''>기존의 여가부도 예산과 권한이 작아 다른 부처와 협력하기 어려웠는데 본부로 격하시킨다면 힘을 더 빼는 것</직접 인용구>이라고 했다.
성평등 정책을 수립하고 정부조직 전반에서 성평등이 지켜지는지 점검할 컨트롤타워가 사라진다는 우려가 나온다.  박선영 한국여성정책연구원 선임연구위원은<직접 인용구 입장=``비판적''>양성평등 정책은 전 부처에 대한 조정 업무이고, 그래서 여성가족부가 만들어졌던 것</직접 인용구>이라며<직접 인용구 입장=``비판적''>집행 기능 중심의 복지부에 집어넣으면 실효성이 떨어질 것</직접 인용구>이라고 지적했다.
기존 여성가족부가 정책 기능을 제대로 수행하지 못했다고 평가하는 전문가들도 기능을 복지부로 축소·이관해선 성평등 정책을 펴기 어렵다고 지적했다. 정재훈 서울여대 사회복지학과 교수는<직접 인용구 입장=``지지적''>성차별 문제와 관련해 그동안 정부에선 여성가족부에서 여성끼리 모여 논의해도 된다는 `성게토화(Gender Ghetto)' 현상이 나타났다</직접 인용구>며<직접 인용구 입장=``지지적''>대통령 직속 성평등위원회를 설치해 성평등 문제를 대통령 의제로 끌어올리는 게 필요하다</직접 인용구>고 지적했다.
홍성걸 국민대 행정학과 교수는<직접 인용구 입장=``지지적''>가족정책의 경우 그동안 복지부의 복지 정책과 분리되면서 정책 동력이 분산되는 부정적인 측면이 있었다</직접 인용구>면서 여가부 개편의 시너지 효과를 긍정적으로 평가했다. 다만 홍 교수도 <직접 인용구 입장=``비판적''>성평등 정책은 전 부처의 모든 정책을 테스트할 수 있는 양성평등위원회를 만들어 맡기는 게  바람직하다</직접 인용구>고 말했다.\\\\
<결론부 입장=``비판적''>여가부 내부에서도 복지부 산하로 들어가면 정책들이 `곁가지 취급' 받을 거라는 우려가 나온다. 한 여가부 관계자는<직접 인용구 입장=``비판적''>여가부의 일은 다양성, 젠더에 대한 인식을 갖고서 스스로 목소리를 낼 수 없는 이들을 보호하는 것</직접 인용구>이라며<직접 인용구 입장=``비판적''>방대한 복지부로 갔을 때 이런 업무의 비중은 미미해질 수밖에 없을 것</직접 인용구>이라고 했다.</결론부>
}}
\end{tcolorbox}
\centering
\vspace*{-4mm}
\caption{The original Korean-language prompt used in \textsc{\mymethod}, shown with an illustrative input. Blue italic text highlights the input.}
\label{fig:proposed_prompt_korean_original}
\end{figure*}

\end{document}